\pdfoutput=1

\documentclass[11pt]{article}

\usepackage[preprint]{acl}

\usepackage{times}
\usepackage{latexsym}
\usepackage{enumitem}

\usepackage[T1]{fontenc}

\usepackage[utf8]{inputenc}

\usepackage{microtype}

\usepackage{inconsolata}

\usepackage{graphicx}
\usepackage{booktabs}   
\usepackage{array}      
\usepackage{url}
\usepackage{ragged2e}
\usepackage{adjustbox}
\usepackage{multirow}
\usepackage{tabularx} 
\usepackage{amsmath}
\usepackage{amssymb}
\usepackage{pifont}
\usepackage{tcolorbox}
\usepackage{geometry}
\usepackage{siunitx}

\usepackage{xcolor}

\title{\textit{Why Stop at One Error?} Benchmarking LLMs as Data Science Code Debuggers for Multi-Hop and Multi-Bug Errors
}

\author{Zhiyu Yang$^{1}$\quad
  Shuo Wang$^{2}$\quad
  Yukun Yan$^{2}$\quad
  Yang Deng$^{1}$\\
  $^1$Singapore Management University  \quad$^2$Tsinghua University \\
  \texttt{kelvin.yangzhiyu@outlook.com, ydeng@smu.edu.sg}
  }

\begin{document}
\maketitle
\begin{abstract}

LLMs are transforming software development, yet most code benchmarks still emphasize syntactic or functional correctness in simple, single-error cases. These settings miss the core difficulty of real-world data science debugging, where runtime errors propagate across multiple lines (multi-hop) and often appear in sets (multi-bug). We introduce \textbf{DSDBench}: \textbf{D}ata \textbf{S}cience \textbf{D}ebugging \textbf{Bench}mark, the first benchmark to systematically evaluate LLMs on this challenge. Unlike general debugging benchmark suites such as SWE-bench, DSDBench targets non-expert, data-centric scripting, where practitioners rely heavily on black-box libraries and write exploratory code that is error-prone and difficult to debug. Evaluations of state-of-the-art LLMs reveal large performance gaps: even frontier models that excel at code generation fail to reliably trace and resolve these errors, exposing a critical “generation versus understanding” gap. DSDBench provides a resource to drive progress toward more robust and trustworthy AI-assisted data science. \footnote{DSDBench is publicly available at \url{https://github.com/KevinCL16/DSDBench}.}

\end{abstract}

\section{Introduction}

Recent advancements in Large Language Models (LLMs) have significantly reshaped software development practices, particularly in automating code generation and debugging. Benchmarks like DebugBench \citep{tian-etal-2024-debugbench}, CodeEditorBench \citep{guo2024codeeditorbenchevaluatingcodeediting}, and DebugEval \citep{yang2025coastenhancingcodedebugging} have played a pivotal role in evaluating LLMs’ capabilities in code repair. However, these benchmarks largely rely on simplified programming exercises from platforms like \textit{LeetCode}, which prioritize \textbf{syntactic correctness} and \textbf{functional accuracy} in \textbf{isolated} and \textbf{single-error} scenarios, far removed from real-world software complexity.

Meanwhile, growing research efforts are exploring LLMs’ potential in data science coding \cite{yang-etal-2024-matplotagent, hu2024infiagentdabenchevaluatingagentsdata, zhang2024benchmarkingdatascienceagents, hong2024datainterpreterllmagent}, where practitioners routinely tackle challenges involving black-box library functions, intricate data transformations, and statistical modeling. Yet, a critical gap persists: despite this emerging focus, there remains a striking lack of investigation into LLMs’ ability to \textit{debug dynamic logical errors in data science code}. Such errors, manifesting only at runtime, are endemic to this domain due to hidden dependencies in data pipelines, implicit assumptions in mathematical operations, and unpredictable interactions with external resources.

\begin{figure*}[t]
    \centering
    \includegraphics[width=0.95\linewidth]{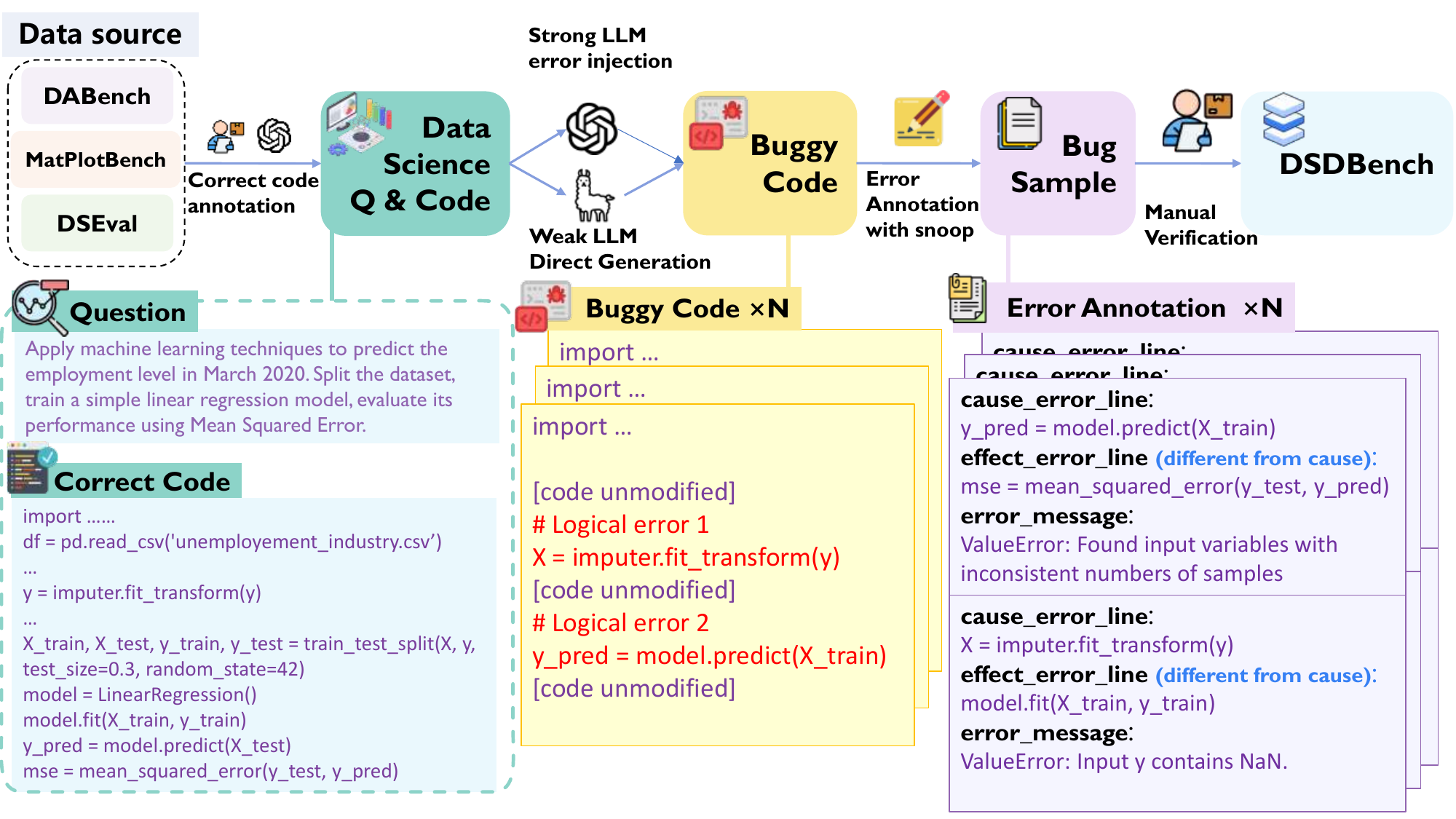}
    \caption{Dataset construction pipeline of DSDBench.}
    \label{fig:workflow}
\end{figure*}

As illustrated in Figure \ref{fig:workflow}, unlike constrained programming exercises, debugging data science codebases presents unique challenges:
1) Its heavy reliance on external libraries (e.g., pandas, NumPy, scikit-learn, matplotlib) means subtle misuses or incorrect data processing steps can easily trigger downstream runtime exceptions. 
2) Data scientists often work in interactive environments like Jupyter Notebooks, which lack robust debugging tools. This makes it harder to identify and fix runtime bugs, especially when \textbf{multiple subtle errors}, such as incorrect data transformations or misaligned indices, coexist and interact within the code, complicating the debugging process.
3) Standard debugging tools offer limited assistance in diagnosing \textbf{multi-hop logical errors} within complex workflows. The root cause of these errors can be distantly located from the point of error manifestation. 
Standard debuggers typically report the \textit{symptom} (the line of error manifestation in the stack trace) rather than the \textit{root cause} responsible for the program's termination. Linking these two is essential not only for reasoning, but also for \textit{trustworthy} assistance grounded in what users actually see.

Overall, a dedicated benchmark for rigorously assessing LLMs' dynamic debugging of multi-hop logical errors in complex multi-bug data science code is still lacking. Unlike general debugging suites such as SWE-bench that evaluate patch-level fixes in professional OSS repositories via unit tests, our setting targets non-expert, data-centric scripting where failures are often data-dependent and surface as interpreter-visible runtime errors. Crucially, we evaluate whether models can connect \textit{root causes} to these interpreter-visible failure points (\textit{effect lines}), a prerequisite for trustworthy explanations.

Motivated by this evident gap in evaluating LLMs' dynamic debugging skills for data science, we introduce \textbf{DSDBench}: the \textbf{D}ata \textbf{S}cience \textbf{D}ebugging \textbf{Bench}mark. 
Distinct from prior works that primarily focus on repairing single, syntactic and static errors, DSDBench is the first benchmark to systematically evaluate LLMs on: (1) \textbf{Multi-Hop Error Tracing}: requiring models to trace runtime errors back through multiple lines of data science code to identify the root cause; and (2) \textbf{Multi-Bug Error Detection}: assessing their ability to concurrently detect and reason about multiple logical errors within a single data science code snippet. We \emph{focus on interpreter-visible (crashing) bugs} to enable scalable, deterministic annotation of cause/effect lines and messages. For clarity, we define a “hop” whenever the cause and effect lines differ, and for multi-bug items a prediction is counted correct only if \emph{all} constituent bugs are identified.
Table \ref{tab:benchmark_comparison_simplified_aesthetics} summarizes the comparisons between DSDBench and existing code debugging benchmarks.

\begin{table}[t]
\small
    \centering
    \scalebox{0.75}{
    \begin{tabular}{lcccc}
        \toprule
        \multirow{2}{*}{\textbf{Benchmark}} & \multirow{2}{*}{\textbf{Domain}} & \textbf{Error} & \textbf{Multi-Hop} & \textbf{Error} \\
        & &\textbf{Complexity} & \textbf{Error} & \textbf{Type}\\
        \midrule
        DebugBench & General & Multi-Bug & \textcolor{red}{\ding{55}} & Static \\
        DebugEval & General & Multi-Bug & \textcolor{red}{\ding{55}} & Static \\
        CodeEditorBench & General & Single-Bug & \textcolor{red}{\ding{55}} & Static \\
        \midrule
        \textbf{DSDBench} & \textbf{Data Science} & Multi-Bug & \textcolor{green}{\ding{51}} & \textbf{Runtime} \\
        \bottomrule
    \end{tabular}
    }
    \caption{Comparison with existing benchmarks.}
    \label{tab:benchmark_comparison_simplified_aesthetics}
\end{table}

DSDBench leverages datasets and tasks from established data science coding benchmarks like DABench \citep{hu2024infiagentdabenchevaluatingagentsdata}, MatPlotBench \citep{yang-etal-2024-matplotagent}, and DSEval \citep{zhang2024benchmarkingdatascienceagents}. We systematically inject errors into data science code, synthesizing multi-error scenarios by combining individual bugs. Our dataset comprises 1,117 meticulously annotated samples, complete with ground-truth cause-effect error line pairs and captured runtime error messages.  

In summary, our contributions are threefold:
\begin{itemize}[leftmargin=*]
    \item \textbf{DSDBench Benchmark:} We release the first dedicated benchmark and dataset for evaluating LLMs in runtime, multi-bug debugging of data science code. DSDBench features realistic logical errors, multi-hop error scenarios, and detailed annotations, addressing a critical gap in current debugging benchmarks.
    \item \textbf{Automated Error Injection and Annotation Framework:} We develop a robust pipeline for automated error injection, runtime execution tracing, and alignment of interpreter outputs with error-originating code lines, facilitating scalable benchmark creation and future expansion.
    \item \textbf{Empirical Analysis and Insights:} We present a comprehensive empirical evaluation of state-of-the-art closed-source and open-source LLMs on DSDBench. Our findings reveal significant performance gaps and highlight critical challenges in dynamic debugging for complex, real-world data science code.
\end{itemize}

\section{DSDBench Construction}
\label{sec:DSDBench_construction}

The creation of a high-quality dataset is paramount for a robust benchmark. As illustrated in Figure \ref{fig:workflow}, DSDBench is meticulously constructed through a multi-stage process encompassing data sourcing, correct code preparation, error injection, error annotation, and quality assurance. 

\begin{table*}[t]
\resizebox{\textwidth}{!}{%
\begin{tabular}{@{}cccccccc@{}}
\toprule
\multicolumn{3}{c}{\textbf{Dataset Size}} & \multicolumn{2}{c}{\textbf{Example Type}} & \textbf{Multi-Error Examples} & \textbf{Code Complexity} & \textbf{Question Complexity} \\ \midrule
Total \# Examples & \# Single-Error & \# Multi-Error & \# Single-hop & \# Multi-hop & Avg Errors/Example & Avg Code Length & Avg Question Length \\ 
1,117 & \multicolumn{1}{c}{741} & 376 & 385 & 356 & 2.87 $\pm$ 1.14 & 65.31 $\pm$ 21.31 & 92.42 $\pm$ 55.86 \\ \bottomrule
\end{tabular}%
}
\caption{Dataset statistics of DSDBench. }
\label{tab:data_stats}
\end{table*}

\subsection{Data Collection}

We build the DSDBench upon three widely-adopted data science coding benchmarks for their realistic data science tasks and diverse scenarios, including DABench \cite{hu2024infiagentdabenchevaluatingagentsdata}, MatPlotBench \cite{yang-etal-2024-matplotagent}, and DSEval \cite{zhang2024benchmarkingdatascienceagents}. 
We focus on the hard subset of DABench because error injection in its easy and medium subsets rarely produces runtime exceptions. MatPlotBench and DSEval supplement DABench, expanding task diversity and library coverage (\texttt{pandas}, \texttt{sklearn}, \texttt{scipy}, \texttt{matplotlib}, \texttt{numpy}) to represent typical data science workflows. These benchmarks cover data manipulation, statistical analysis, machine learning, and visualization.  


However, some of these datasets mainly contain the natural language instructions and the final results after running the data science code, while the ground-truth correct codes are not provided. As the first step, we prepare the correct and error-free codes for each benchmark as follows: 

\paragraph{DABench} We design an agent-based annotation framework, which includes a self-debugging code agent and an error verifier agent. Annotation begins by feeding benchmark questions and metadata to the self-debugging code agent, which generates initial code and debugs it based on error messages. Subsequently, the error verifier agent analyzes this code to correct logical errors, meanwhile ensures the code produces correct answers according to DABench's ground truths. The details of the agent-based annotation framework are presented in Appendix \ref{app: data_annotation}. 

\paragraph{MatPlotBench} Similar agent-based code generation is adopted, but automated verification is challenging due to the visual nature of plot outputs. Therefore, manual expert verification is employed, comparing plots to ground truth images and correcting code for accurate visualizations. 

\paragraph{DSEval} We extract and concatenate code blocks from ground truth Jupyter notebooks provided by DSEval, using concatenated code as our benchmark's correct code. 

\subsection{Error Injection}
\label{subsubsection:error_injection}

To systematically introduce errors, we employ two complementary methods. This dual design ensures both realism and diversity of bug types. The details of prompts are provided in Appendix \ref{app:error_injection}.

\paragraph{Strong LLM-based Error Injection}
We instruct GPT-4o to inject runtime-interrupting bugs into otherwise correct solutions. Step 1: identify lines invoking core data science libraries (\texttt{numpy}, \texttt{scipy}, \texttt{matplotlib}, \texttt{sklearn}, \texttt{pandas}). Step 2: modify them to introduce plausible faults such as API misuse, incorrect parameter settings, or data-shape mismatches (e.g., NaNs, inconsistent dimensions). These produce realistic logic/runtime bugs. Importantly, injection often yields \textbf{multi-hop errors}: the causal bug occurs earlier in the program, but the crash manifests later (e.g., a faulty imputation call causing a failure during model training).

\paragraph{Weak LLM-based Direct Error Generation}
We also instruct Llama-3.1-8B to write entire solutions from scratch. Due to its weaker capability, the generated code frequently contains beginner-style mistakes (e.g., missing imports, wrong API signatures, incomplete function definitions). These naturally introduce \textbf{multi-hop errors}, for example when a sub-function mistake surfaces as an exception in the \texttt{main} function.

\paragraph{Rationale} Using both approaches diversifies error characteristics: strong LLM injection yields semantically rich runtime bugs, while weak LLM generation reflects novice-style mistakes. This mitigates bias from relying on a single method.




\subsection{Error Annotation}
\label{subsubsection:annotation}

For each buggy snippet, we annotate three ground truths: \texttt{cause\_error\_line}, \texttt{effect\_error\_line}, and \texttt{runtime error message}. We then construct both single-error and multi-error cases, requiring models to reason across causal and manifested locations.

\paragraph{Single-Error Annotation}
\label{subsubsection:single_error_annotation}
We commence dynamic error capture with \texttt{snoop}\footnote{\url{https://pypi.org/project/snoop/}}, a Python debugging library that logs execution details, for single-error ground truth.  \texttt{snoop} monitors the execution of both injected and direct generated error code.  We first filter out successfully executed ones. For error-triggering snippets, we analyze \texttt{snoop}'s execution traces to extract: \texttt{cause\_error\_line} (error origin), \texttt{effect\_error\_line} (error manifestation), and \texttt{runtime error messages}, providing ground truths for single-error annotation.

\paragraph{Multi-Error Annotation}
\label{subsubsection:multi_error_annotation}
We create multi-error cases by systematically combining validated single errors into candidate pools per question, from which random subsets are sampled. An instance is marked \emph{correct} only if \emph{all} constituent bugs are identified. Although each bug is logically independent, runtime interactions (e.g., early crashes masking later ones) can increase difficulty, which we preserve to reflect realistic debugging.

\paragraph{Multi-Hop Definition}
We define a ``hop'' as the spatial separation between the cause line (where the bug originates) and the effect line (where execution fails). The hop distance may be a few lines within the same function (no change in call stack) or span across functions (involving call stack depth). The essential characteristic is that the interpreter reports a crash at a different line than where the bug was introduced, forcing models to trace execution flow.

\subsection{Human Verification and Quality Control}
\label{subsubsection:human_verification}

To ensure the quality and correctness of the constructed dataset, we perform a two-stage verification process: \textbf{code-based checks} and \textbf{LLM-assisted verification}. 1) Code checks involve printing and manually inspecting annotated cause and effect lines to correct nonsensical annotations by human annotators. We also print error messages, identifying and resolving a common \texttt{plt.show()} backend issue by adding backend settings to the MatPlotBench correct code examples. 2) LLM-assisted verification is used to review all annotations, flagging remaining inconsistencies that require human intervention to correct the annotations. 
Overall, the pass rates of the human verification for the two stages are 83\% and 87\%, respectively. The high pass rates also validate the effectiveness of the automated annotation process.

\paragraph{Bug Distribution, Realism, and Scope}
We approximate non-expert data-science scripting, where code is exploratory and loosely structured. To mitigate LLM-specific biases, our injection is constrained and modular: a strong LLM first identifies API-invoking lines and then performs \emph{atomic, plausible} edits (e.g., argument name/value mistakes, missing \texttt{axis}, shape mismatches), limiting unrealistic global rewrites and better mimicking small mistakes that propagate during execution. Weak-LLM generation complements this by introducing novice-style patterns at the program level. We presently \emph{focus on interpreter-visible (crashing) bugs} to support reliable, scalable labeling; functional bugs without hard crashes are deferred to future work. Empirically, models with stronger coding/reasoning (e.g., Claude-3.5-Sonnet, DeepSeek-V3, LRMs) often outperform GPT-4o (used for injection), suggesting DSDBench rewards genuine reasoning rather than model-specific shortcuts.

\subsection{Dataset Characteristics}
\label{subsec:dataset_characteristics}

This section presents a statistical overview of the DSDBench dataset, characterizing its composition, diversity, and complexity. Table \ref{tab:data_stats} provides a statistical overview of the DSDBench dataset. The \textbf{dataset size and splits} are as follows: the total number of examples is 1,117, of which 741 are single-error examples and 376 are multi-error examples. For single-error examples, the number of examples with multi-hop cause and effect error lines is 356, the rest 385 examples contain identical cause and effect error lines \textit{i.e.}, single-hop errors). For multi-error examples, the number of errors per example ranges from 2 to 9, with an average of 2.87 errors per example. Regarding complexity, the average code length is 65.31 lines, and the average question length is 92.42 words. For a more detailed breakdown of these statistics, please refer to Table \ref{tab:data_stats}.

\begin{figure}[t]
    \centering
    \includegraphics[width=0.99\columnwidth]{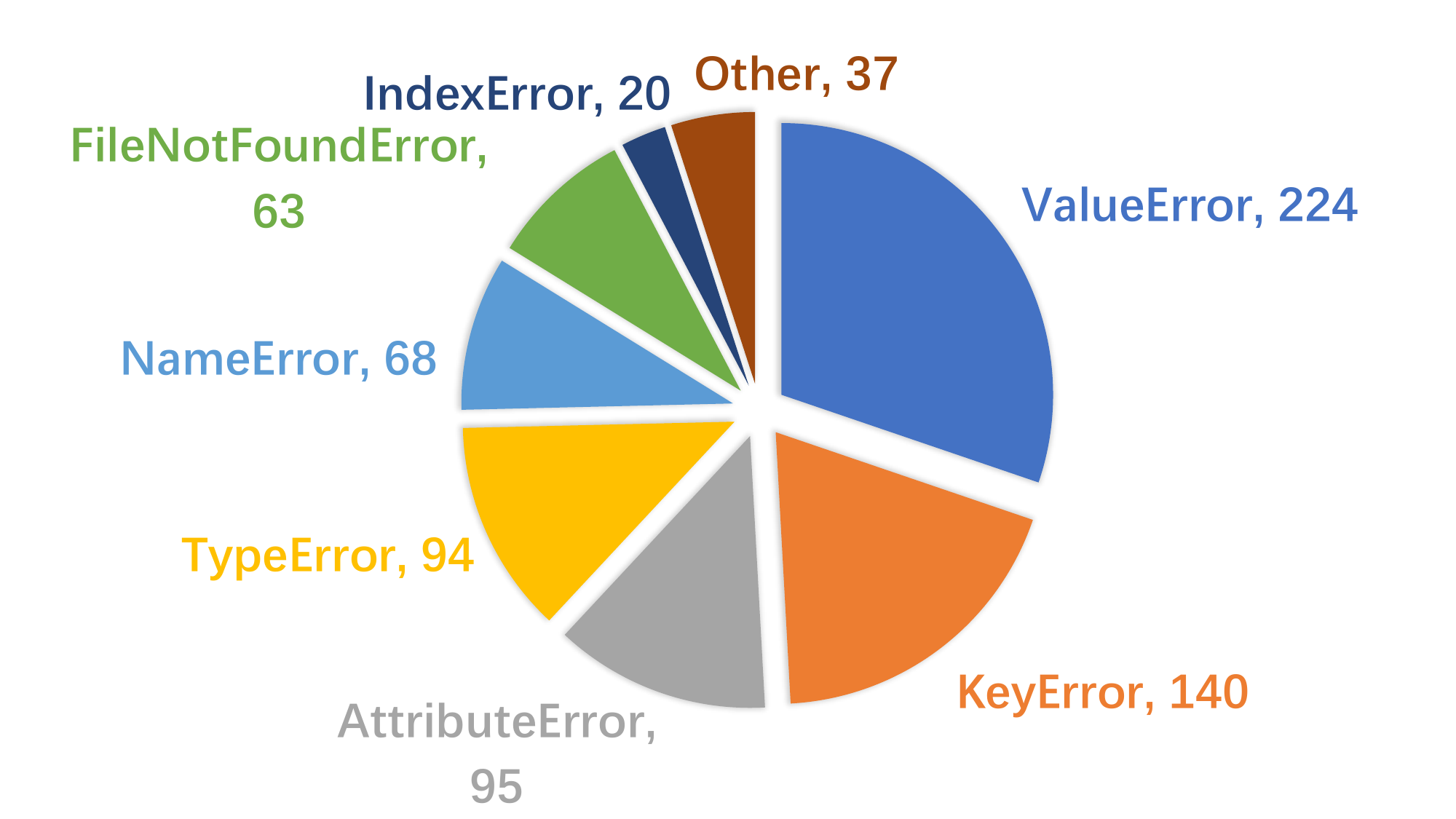}
    \caption{Distribution of different error types. Details of error types are described in Appendix \ref{app:error_type}.}
    \label{fig:error_type}
\end{figure}

\begin{figure}[t]
    \centering
    \includegraphics[width=0.99\columnwidth]{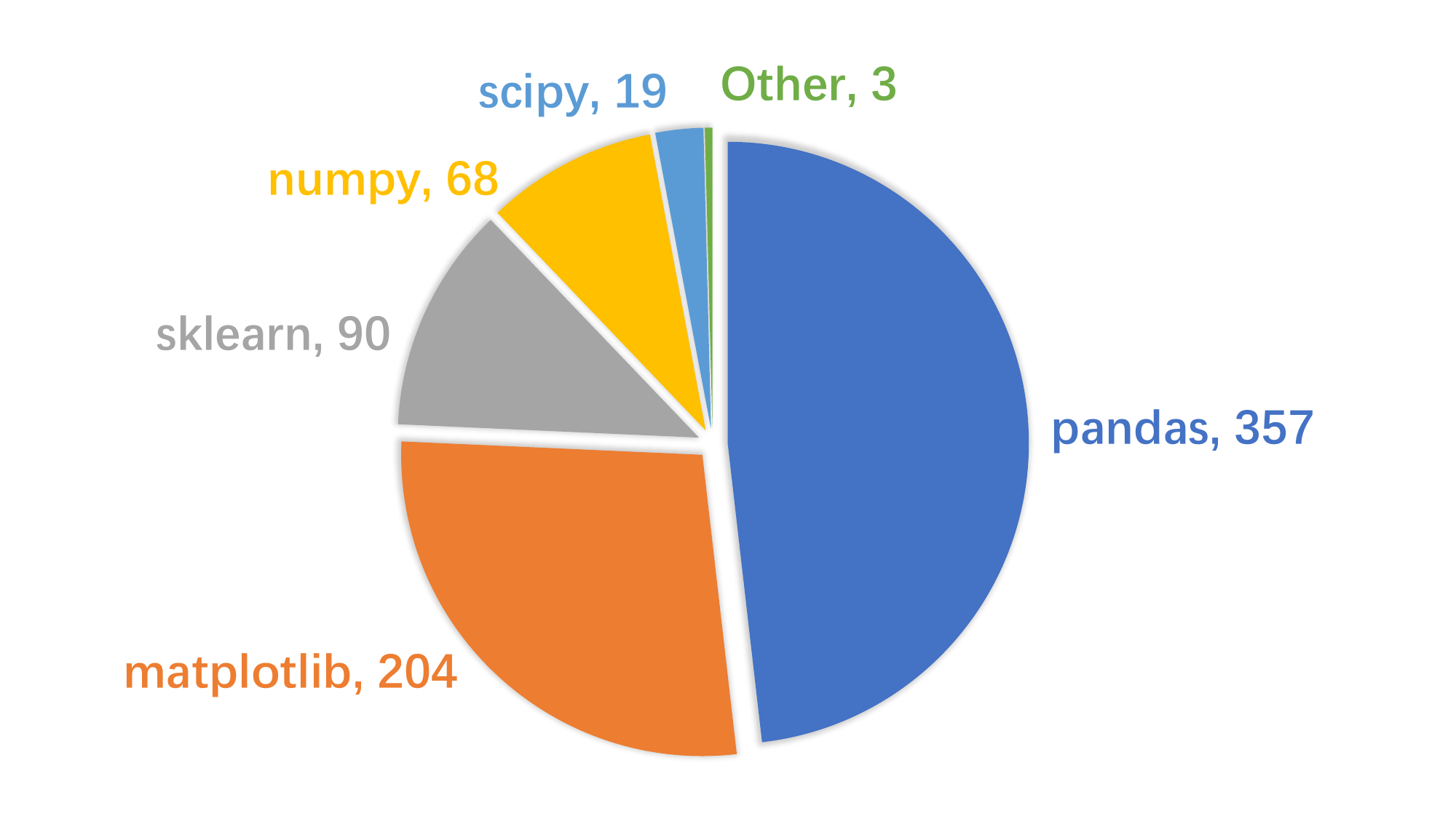}
    \caption{Distribution of different data science libraries.}
    \label{fig:library_usage}
\end{figure}

Figure \ref{fig:error_type} summarizes the \textbf{error type distribution}, and Figure \ref{fig:library_usage} shows the \textbf{library coverage}. Together they indicate broad coverage across common runtime/data-driven failures and core APIs, lending distributional realism despite synthetic injection.

\section{Problem Formulation}

\subsection{Task Definition}
\label{subsec:task_definition}

This section formally defines the task of \textbf{Data Science Code Debugging} for
the DSDBench benchmark, outlining the input, desired output, and evaluation settings. The primary objective of DSDBench is to evaluate the capability of LLM-based debuggers to identify and explain \textbf{logical errors} in \textbf{data science Python code} during simulated \textbf{runtime execution}.

The benchmark targets two critical dimensions: \textbf{multi-hop error detection} and \textbf{multi-bug error detection}. \textbf{Multi-hop} evaluates whether models can trace errors to their \textbf{root cause (\texttt{cause\_error\_line})} when it differs from the \textbf{interpreter's error point (\texttt{effect\_error\_line})}. We define a ``hop'' whenever these two lines are not identical; the distance may be within the same function (no call-stack change) or across function calls (with call-stack depth). This spatial separation forces reasoning over execution flow. \textbf{Multi-bug} requires identifying and explaining \emph{all} concurrent logical errors in a snippet, not merely the first encountered, mirroring real debugging where missing one error leaves the program incorrect. We also evaluate \textbf{error message reproduction}, i.e., whether models can semantically reproduce the \textbf{interpreter-thrown messages} for each identified error.

Formally, for each task instance $i$, the input is a pair $(Q_i, C_i)$, where $Q_i$ is a natural language question describing a data science task, and $C_i$ is a Python code snippet intended to perform task $Q_i$, but containing logical errors. The task of the LLM is to predict a structured output $O_i = (L_{cause, i}, L_{effect, i}, M_i)$, where $L_{cause, i}$ is the exact line of code for the cause error, $L_{effect, i}$ is the exact line of code for the effect error, and $M_i$ is the error message that would be produced by a Python interpreter when executing $C_i$. The DSDBench benchmark dataset can be represented as $D = \{(Q_i, C_i, L_{cause, i}^{GT}, L_{effect, i}^{GT}, M_i^{GT})\}_{i=1}^{N}$, where $GT$ denotes the ground truth annotation.  The objective is to evaluate LLMs' capabilities to perform the task of $f: (Q_i, C_i) \mapsto O_i$ which localizes and interprets the error.


\subsection{Evaluation Metrics}
\label{subsec:evaluation_metrics}

This section details evaluation metrics for LLM debugger performance on DSDBench, focusing on error localization accuracy and description quality.  
Model performance is evaluated across four dimensions, including \textbf{Cause Line Matching}, \textbf{Effect Line Matching}, \textbf{Error Type Matching}, and \textbf{Error Message Matching}.

We compute \texttt{cause\_line\_score}, \texttt{effect\_line\_score}, and \texttt{error\_type\_score} as binary exact matches (1 for exact match with ground truth, 0 otherwise). For error messages, exact string matching is too brittle: LLMs may produce semantically equivalent messages with different surface forms. We therefore adopt a semantic rubric scored by GPT-4o on a five-point Likert scale \{0.0, 0.25, 0.5, 0.75, 1.0\}. Example: \emph{GT:} ``ValueError: Input contains NaN'' vs. \emph{Prediction:} ``ValueError: The model cannot handle NaN values in the input''—lexically different but semantically equivalent (scored $\ge 0.75$). The full rubric and additional examples are provided in Appendix~\ref{app:evaluation_prompt}.



\paragraph{Dimension-Level Definitions:} For each evaluated dimension:
\begin{itemize}[leftmargin=*,nosep]
    \item \textbf{TP (True Positives):} Number of instances with correct LLM predictions (exact match for lines/types, \texttt{error\_message\_score} $\ge$ 0.75 for error messages).
    \item \textbf{FP (False Positives):} Number of instances with specific incorrect LLM predictions (commission errors).
    \item \textbf{FN (False Negatives):} Number of instances where LLM failed to provide a relevant prediction, (omission errors) e.g., incorrect output format ; \(FN = \textit{GT\_Instances} - (TP + FP)\).
    \item \textbf{\textit{GT\_Instances}:} Total Ground Truth Instances for the dimension.
\end{itemize}

\paragraph{Evaluation Metrics (per dimension):}
We employ Precision, Recall, F1-score, and Accuracy to evaluate performance across dimensions. Because DSDBench only contains test cases with errors, meaning there is no True Negatives in model predictions. Therefore, we calculate \textbf{Recall} by \textbf{(True Positive Rate - TPR)} to measure the completeness of error detection as:
\[ \text{Recall (TPR)} = \frac{TP}{\textit{GT\_Instances}} \]
making Recall (TPR) numerically equivalent to Accuracy. All metrics are calculated dimension-wise to provide a detailed performance profile.

\begin{table*}[t]
    \centering
    
    \begin{adjustbox}{width=\textwidth}
    \begin{tabular}{l|cc|cc|cc|cc}
        \toprule
         \multirow{2}{*}{\textbf{Model}} & \multicolumn{2}{c|}{\textbf{Cause Line}} & \multicolumn{2}{c|}{\textbf{Effect Line}} & \multicolumn{2}{c|}{\textbf{Error Type}} & \multicolumn{2}{c}{\textbf{Error Message}} \\
         & \textbf{Single-Bug} & \textbf{Multi-Bug} & \textbf{Single-Bug} & \textbf{Multi-Bug} & \textbf{Single-Bug} & \textbf{Multi-Bug} & \textbf{Single-Bug} & \textbf{Multi-Bug} \\
        \midrule
         GPT-4o & 39.0 & 20.3 & 34.3 & 10.4 & 30.6 & \textbf{3.6} & 31.4 & \textbf{4.7} \\
         GPT-4o-mini & 40.2 & 11.2 & 23.9 & 2.7 & 21.7 & 2.2 & 21.3 & 0.8 \\
         Claude 3.5 sonnet & 43.7 & 12.3 & 35.2 & 4.1 & \textbf{36.3} & 1.9 & 34.0 & 2.5 \\
         Deepseek-V3 & \textbf{48.3} & 15.1 & 34.5 & 6.6 & 35.9 & 3.3 & \textbf{34.7} & \textbf{4.7} \\
         Llama-3.1-8B-instruct & 25.2 & 3.0 & 14.2 & 0.0 & 7.7 & 0.0 & 7.2 & 0.0 \\
         Llama-3.1-70B-instruct & 42.5 & 0.0 & 29.3 & 0.0 & 20.4 & 0.0 & 20.9 & 0.0 \\
         Llama-3.1-405B-instruct & 41.7 & 18.6 & 31.3 & 8.5 & 29.3 & 1.1 & 29.3 & 2.5 \\
         Qwen2.5-7B-Instruct & 29.3 & 4.7 & 19.3 & 1.1 & 10.7 & 0.3 & 10.9 & 0.0 \\
         Qwen2.5-32B-Instruct & 40.9 & 17.5 & 30.5 & 6.3 & 24.7 & 2.2 & 24.7 & 2.2 \\
         Qwen2.5-72B-Instruct & 41.6 & \textbf{21.4} & \textbf{36.2} & \textbf{11.2} & 27.5 & 3.0 & 27.4 & 3.6 \\
        \bottomrule
    \end{tabular}
    \end{adjustbox}
    \caption{Overall evaluation results of LLMs on DSDBench. The reported score is the Accuracy (\%), while full metrics are presented in Appendix \ref{app:full_metrics}.}\label{tab:combined_bug_accuracy_percent_side_by_side_vertical_lines}
\end{table*}

\begin{table*}[t]
    \centering
    
    \begin{adjustbox}{width=\textwidth}
    \begin{tabular}{ll|cc|cc|cc|cc}
        \toprule
         & \multirow{2}{*}{\textbf{Model}} & \multicolumn{2}{c|}{\textbf{Cause Line}} & \multicolumn{2}{c|}{\textbf{Effect Line}} & \multicolumn{2}{c|}{\textbf{Error Type}} & \multicolumn{2}{c}{\textbf{Error Message}} \\
        & & \textbf{Single-Bug} & \textbf{Multi-Bug} & \textbf{Single-Bug} & \textbf{Multi-Bug} & \textbf{Single-Bug} & \textbf{Multi-Bug} & \textbf{Single-Bug} & \textbf{Multi-Bug} \\
        \midrule
        \multirow{3}{*}{\textbf{LLMs}} & GPT-4o & 35.4 & 12.5 & 31.2 & 5.0 & 33.3 & 2.5 & 33.3 & 2.5 \\
        & GPT-4o-mini & 39.6 & 7.5 & 29.2 & 5.0 & 25.0 & 2.5 & 22.9 & 0.0 \\
        & Deepseek-V3 & \textbf{44.8} & 12.5 & 28.1 & 7.5 & \textbf{34.4} & 5.0 & \textbf{34.4} & 7.5 \\
        \midrule
        \multirow{3}{*}{\textbf{LRMs}} & Gemini 2.0 Flash Thinking & 42.7 & 20.0 & \textbf{32.3} & 12.5 & 33.3 & 0.0 & 35.4 & 2.5 \\
        & Deepseek-R1 & \textbf{49.0} & 32.5 & \textbf{49.0} & \textbf{25.0} & \textbf{53.1} & \textbf{15.0} & \textbf{54.2} & \textbf{17.5} \\
        & o1-mini & 43.8 & \textbf{35.0} & 36.5 & 22.5 & 43.8 & \textbf{17.5} & 46.9 & \textbf{17.5} \\
        \bottomrule
    \end{tabular}
    \end{adjustbox}
    \caption{Comparison with large reasoning models (LRMs). The reported score is the Accuracy (\%), while full metrics are presented in Appendix \ref{app:full_metrics}. Due to the unstableness of certain LRM APIs, we randomly sample a subset of DSDBench for this evaluation, which comprises of 96 Single-Error and 40 Multi-Error instances.}\label{tab:combined_bug_accuracy_subset_percent_side_by_side_vertical_lines}
\end{table*}

\section{Experiments}

\subsection{Setup}
\label{subsec:experiment_setup}

\paragraph{Models}
We benchmarked a diverse set of state-of-the-art models on the DSDBench dataset, including both closed-source models and open-source models.  Specifically, the closed-source models we employed were GPT-4o, GPT-4o-mini, o1-mini \citep{openai2024openaio1card}, Gemini 2.0 Flash Thinking \citep{geminiteam2024geminifamilyhighlycapable}, and Claude 3.5 sonnet-20240620. Open-source model consisted of Llama-3.1-8B-instruct, Llama-3.1-70B-instruct, Llama-3.1-405B-instruct \citep{grattafiori2024llama3herdmodels}, Qwen2.5-7B-Instruct, Qwen2.5-32B-Instruct, Qwen2.5-72B-Instruct \citep{qwen2025qwen25technicalreport}, DeepSeek-V3, DeepSeek-R1 \citep{deepseekai2025deepseekr1incentivizingreasoningcapability}.  Notably, we categorize Gemini 2.0 Flash Thinking, DeepSeek-R1 and o1-mini as Large Reasoning Models (LRMs). All models were used with their default decoding parameters apart from setting temperature to 0. Zero-shot setting were used. We used OpenRouter's API services for all models.

\paragraph{Evaluation Protocol}

The evaluation prompt is identical across models and includes the task description, the buggy DSDBench Python snippet, and instructions to output a structured JSON diagnosis. We \emph{disallow external tools and execution} to isolate in-context reasoning. For \emph{multi-bug} items, a prediction is counted correct only if \emph{all} constituent bugs are identified (no partial credit), consistent with Section~\ref{subsubsection:multi_error_annotation}. The template appears in Appendix~\ref{app:evaluation_prompt}. Metrics follow Section~\ref{subsec:evaluation_metrics}.

A pilot study with agentic systems is reported in Appendix~\ref{app:agentic_pilot}; here we focus the main results on in-context localization without execution.

\subsection{Main Results}
\label{subsec:main_results}

Table~\ref{tab:combined_bug_accuracy_percent_side_by_side_vertical_lines} and Table~\ref{tab:combined_bug_accuracy_subset_percent_side_by_side_vertical_lines} present the primary results of our experiments, showing the accuracy of various models in detecting single and multi-bug scenarios across the full and subset DSDBench datasets.

\paragraph{Single-Bug Debugging Performance}
As shown in Tables \ref{tab:combined_bug_accuracy_percent_side_by_side_vertical_lines}, top-performing LLMs like Deepseek-V3 and Claude 3.5 sonnet achieve reasonable accuracy across all tasks, indicating a degree of error tracing capability.  Conversely, smaller models such as Llama-3.1-8B-instruct and Qwen2.5-7B-Instruct exhibit significantly lower accuracy. Notably, Qwen2.5-72B-instruct demonstrated strong performance, on par with state-of-the-art closed-source LLMs such as GPT-4o and Claude 3.5 sonnet. In general, effect line accuracy is consistently lower than cause line accuracy across models, showing LLMs' deficiency to reason about code execution traces and find the exact location where the program would trigger an error.  Error type and error message accuracy vary across different models, suggesting varying levels of understanding and interpretation of runtime errors.

\paragraph{Challenges in Multi-Bug Debugging}
The results reveal a dramatic decrease in accuracy when models are challenged with multi-bug scenarios, models fails to identify an correct set of errors within a code snippet with multiple bugs. Even for the best-performing models, cause line accuracy drops to around 20\% on the full dataset and 30\% on the subset. This substantial performance degradation underscores the increased complexity of debugging multiple bugs concurrently.  Furthermore, the low accuracy in error type and error message prediction in multi-bug cases suggests that models struggle to correctly interpret error messages within these more complex contexts.

\paragraph{LRMs Show Promise in Multi-Bug Debugging}
Comparing LLMs and LRMs on the subset dataset (Table \ref{tab:combined_bug_accuracy_subset_percent_side_by_side_vertical_lines}) reveals that LRMs generally outperform standard LLMs, particularly in the more demanding multi-bug scenarios, indicating superior reasoning capabilities in LRMs are crucial for tackling complex debugging tasks. A more detailed analysis and case study can be found in Figure \ref{fig:case}.

\begin{figure}[t]
    \centering
    \includegraphics[width=0.99\columnwidth]{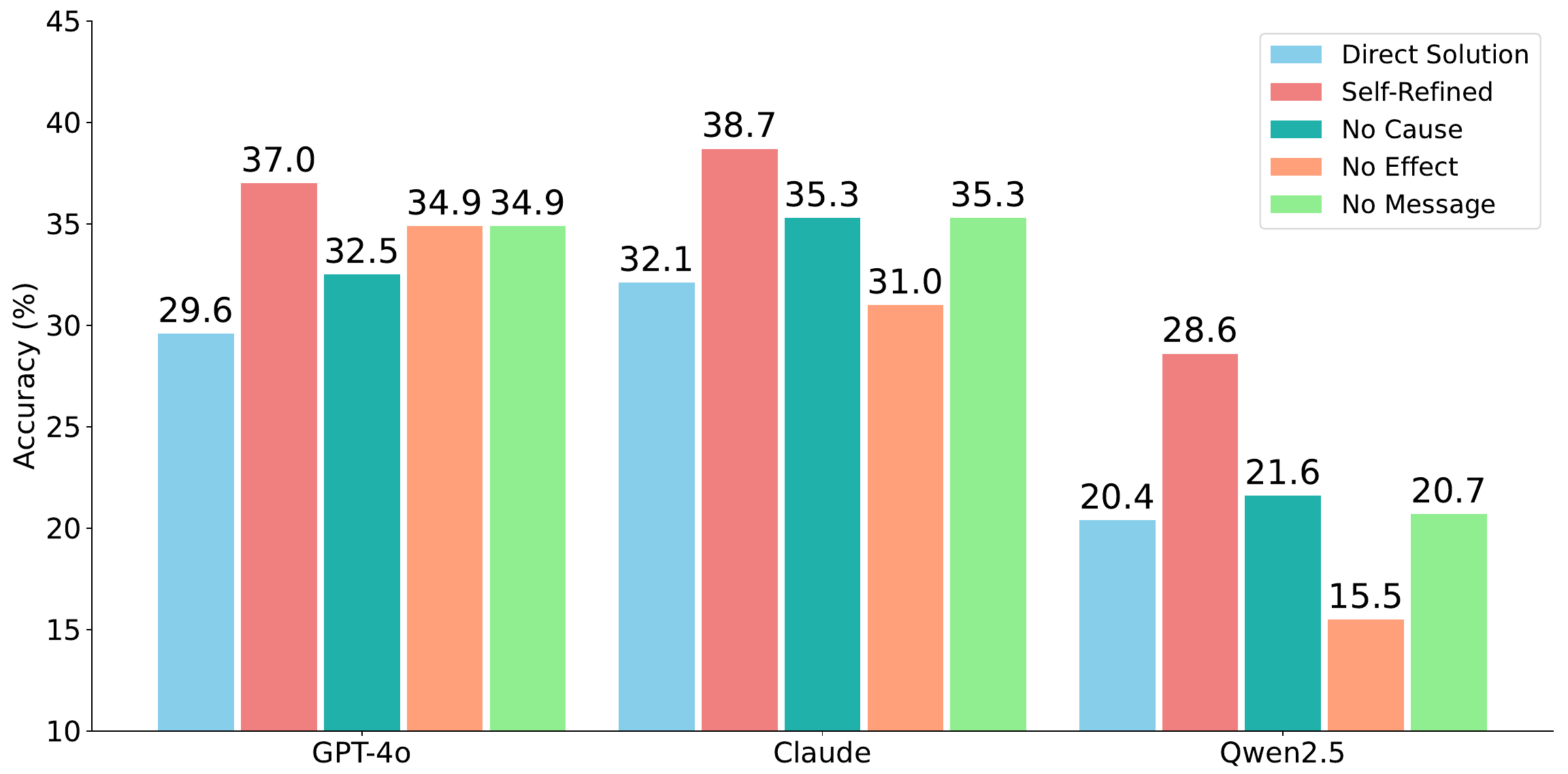}
    \caption{Impact of Self-Refinement.}
    \label{fig:self-refine}
\end{figure}

\subsection{Impact of Self-Debugging}
\label{subsec:self_refinement}

To study how DSDBench-style diagnosis helps downstream coding, we use models’ own debugging outputs as guidance to solve DABench-Hard, comparing direct solutions vs. self-refined solutions that consume \emph{(cause line, effect line, message)}.

Figure \ref{fig:self-refine} shows that \textbf{self-refinement} improves accuracy across GPT-4o, Claude 3.5 Sonnet, and Qwen2.5-72B. Ablations confirm that \emph{lines} carry most of the value—removing \textit{Cause}/\textit{Effect} hurts more than removing \textit{Message}.

\begin{table}[t]
\centering
\resizebox{0.99\columnwidth}{!}{ 
\begin{tabular}{lcccccc}
\toprule
Error Type & \multicolumn{3}{c}{Cause Line} & \multicolumn{3}{c}{Effect Line} \\
\cmidrule(lr){2-4} \cmidrule(lr){5-7}
& GPT-4o & Qwen & DeepSeek & GPT-4o & Qwen & DeepSeek \\
\midrule
ValueError & 57.9 & 61.6 & \textbf{66.1} & 50.5 & \textbf{59.6} & 54.0 \\
TypeError & 30.8 & 39.5 & \textbf{50.0} & 31.9 & 34.6 & \textbf{37.8} \\
NameError & 68.2 & 64.0 & \textbf{85.4} & 56.1 & \textbf{60.0} & 52.1 \\
KeyError & 22.7 & 28.4 & \textbf{37.8} & \textbf{22.7} & 17.6 & 27.9 \\
AttributeError & 35.1 & \textbf{40.5} & 40.0 & \textbf{22.3} & 14.9 & 15.0 \\
IndexError & 36.8 & \textbf{41.2} & 38.9 & 36.8 & \textbf{58.8} & 55.6 \\
FileNotFoundError & 0.0 & 9.6 & \textbf{13.0} & 1.6 & 9.6 & \textbf{11.1} \\
Other & 38.5 & 53.3 & \textbf{66.7} & 23.1 & \textbf{46.7} & 33.3 \\
\bottomrule
\end{tabular}
} 
\caption{Precision w.r.t. different error types. The \textbf{bold} scores represent the best model performance across error types and prediction tasks.}
\label{tab:precision_error_type}
\end{table}

\begin{table}[t]
\centering
\resizebox{0.99\columnwidth}{!}{ 
\begin{tabular}{lcccccc}
\toprule
Library & \multicolumn{3}{c}{Cause Line} & \multicolumn{3}{c}{Effect Line} \\
\cmidrule(lr){2-4} \cmidrule(lr){5-7}
& GPT-4o & Qwen & DeepSeek & GPT-4o & Qwen & DeepSeek \\
\midrule
matplotlib & 46.6 & 48.4 & \textbf{55.6} & 45.6 & 52.2 & \textbf{55.6} \\
numpy & 41.4 & 40.4 & \textbf{44.0} & \textbf{37.9} & 36.8 & 32.0 \\
pandas & 28.1 & 37.0 & \textbf{41.0} & 21.6 & 22.0 & \textbf{24.3} \\
sklearn & 65.1 & 72.5 & \textbf{87.7} & 58.1 & \textbf{63.8} & 53.8 \\
scipy & 36.4 & 54.5 & \textbf{72.7} & 18.2 & 36.4 & \textbf{45.5} \\
\bottomrule
\end{tabular}
} 
\caption{Precision w.r.t. different libraries. }
\label{tab:library_precision}
\end{table}

\subsection{Detailed Analysis}

This section analyzes model performance across error types, libraries, error counts, and multi-hop/single-hop to identify strengths and weaknesses. We adopt GPT-4o, Qwen-72B-Instruct, and DeepSeek-V3 for analysis.

\paragraph{Performance by error types}

Table \ref{tab:precision_error_type} shows error type precision. Models exhibit varying performance on different error types. Generally, models perform better on more common error types and less on more obscure error types. Low performance on FileNotFoundError is possibly attributed to models not having access to the coding environment and file system. DeepSeek-V3 performs best on identifying Cause Lines, scoring the highest on every error type except AttributeError and IndexError. Qwen-72B-Instruct performs best on identifying Effect Lines.

\begin{figure}[t]
    \centering
    \includegraphics[width=0.99\columnwidth]{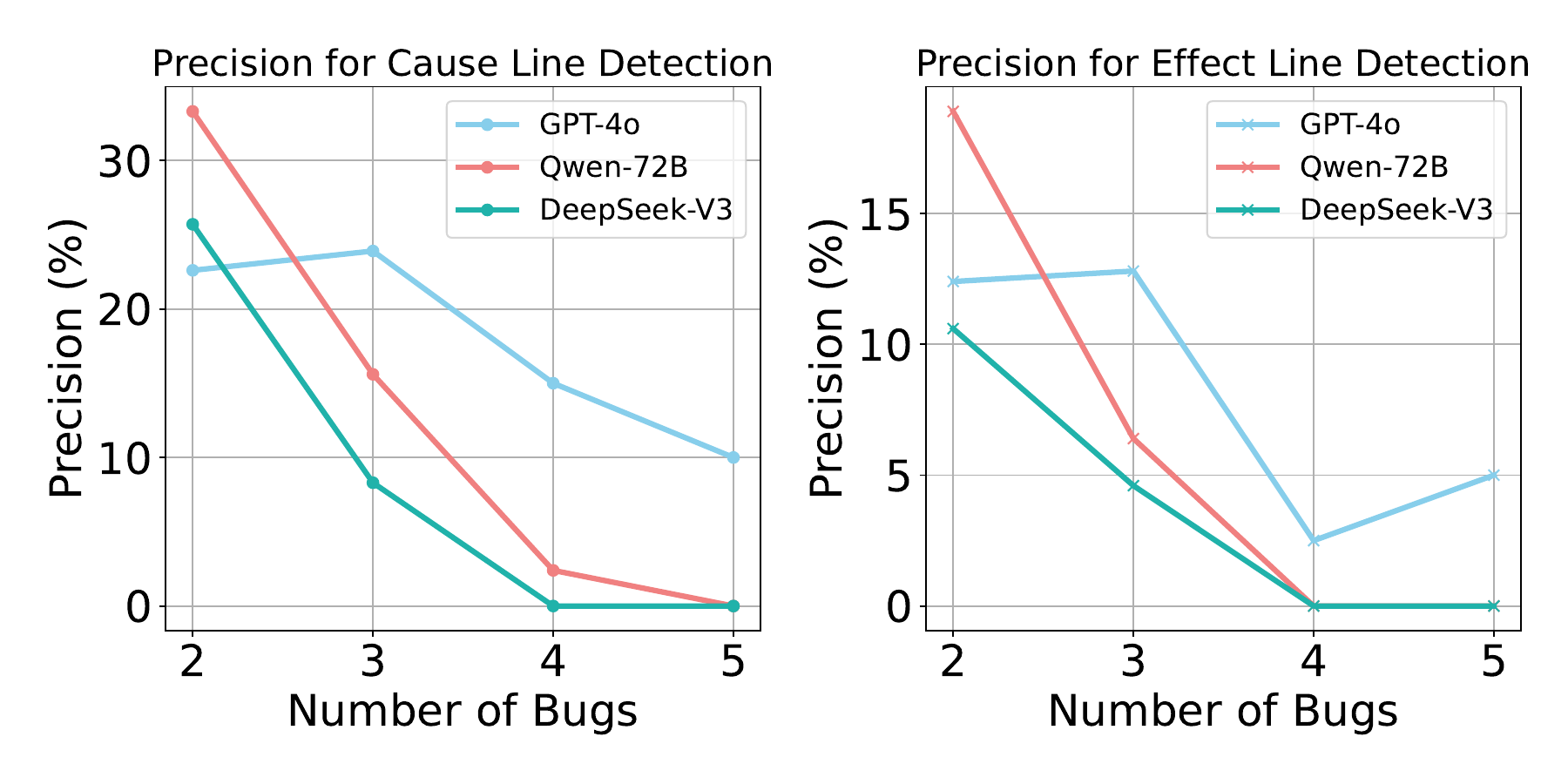}
    \caption{Precision for multi-Bug detection with different number of errors.}
    \label{fig:bug_number_scaling}
\end{figure}

\begin{figure}[t]
    \centering
    \includegraphics[width=0.99\columnwidth]{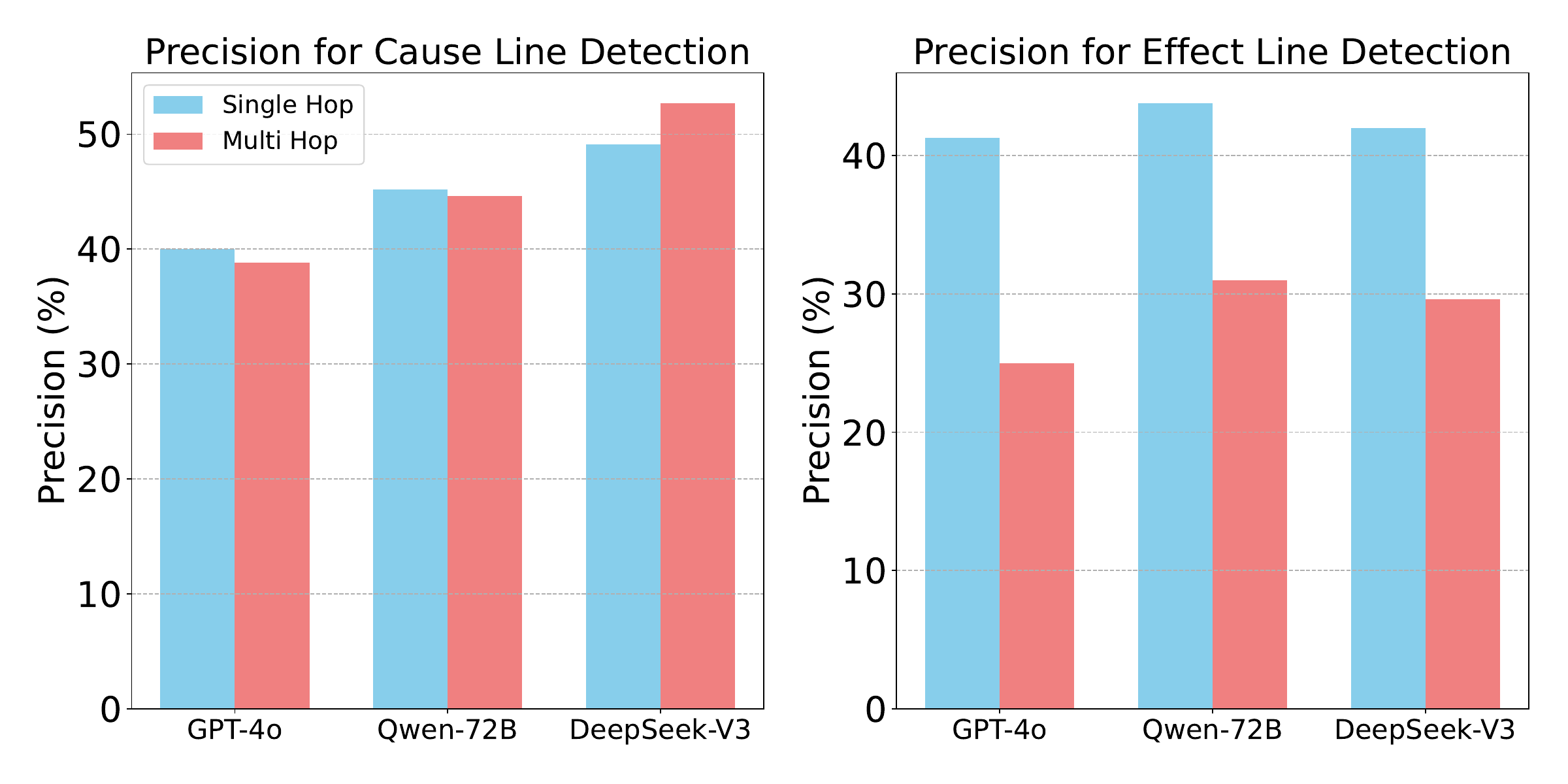}
    \caption{Precision for single-bug detection comparing multi-hop and single-hop errors.}
    \label{fig:perf_single_multi_hop}
\end{figure}

\begin{figure*}[t]
    \centering
    \includegraphics[width=0.95\linewidth]{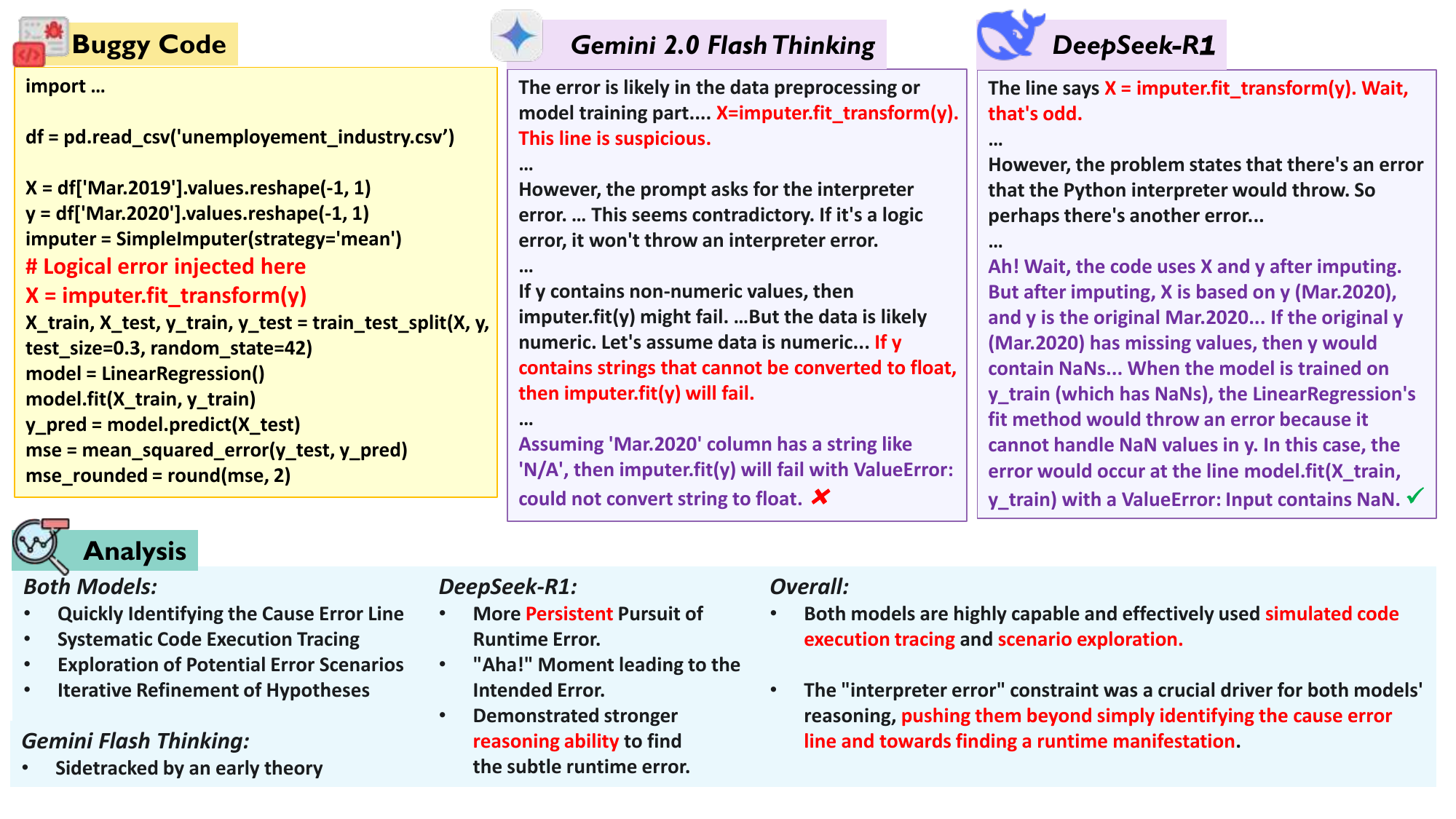}
    \caption{Case study of LRMs.}
    \label{fig:case}
\end{figure*}

\paragraph{Performance by data science libraries}

Table \ref{tab:library_precision} shows library-specific precision. Pandas is the most difficult library to debug, due to its intricate and black-box data manipulation. Models demonstrated best performance on scikit-learn and reasonable performance on matplotlib, numpy and scipy, with significant room for improvement. DeepSeek-V3 demonstrated consistently better performance than other models across all libraries, showing robustness on all kinds of data science coding tasks.

\paragraph{Performance by number of errors}

Figure \ref{fig:bug_number_scaling} shows precision by error count. Precision decreases significantly with more errors for all models. Precision drops sharply when there exists more than 3 errors, highlighting difficulty in multi-error scenarios. GPT-4o performs best on retaining precision on more than 2 bugs, showing potential in concurrent debugging.

\paragraph{Performance on single-hop vs. multi-hop errors}

Figure \ref{fig:perf_single_multi_hop} compares single-hop vs. multi-hop error precision. Cause Line precision is similar for both types of errors across models. Effect Line precision is significantly lower for multi-hop errors, indicating increased difficulty when locating the point where the program triggers an exception. DeepSeek-V3 demonstrated best performance on Cause Line detection on both single-hop and multi-hop errors, while Qwen2.5-72B-Instruct performs slightly better on Effect Line detection.

\paragraph{Prompting Strategies}
We additionally benchmark several test-time prompting strategies (Zero-shot/One-shot CoT, Self-Refine, Self-Consistency) on a single-bug subset. Self-Consistency yields the largest gains for some models, but overall performance remains far from saturation. Full setup and results appear in Appendix~\ref{app:prompting_strategies}.

\paragraph{Agentic Systems}
We also ran a pilot with agentic coding systems (Cursor Agent with Claude 3.5 Sonnet, and OpenHands). Single-attempt repair boosts performance over standalone localization, while unconstrained workflows can approach 100\% pass rate by broadly rewriting code. Full setup, tables, and analysis are provided in Appendix~\ref{app:agentic_pilot}.

\subsection{Case Study}

According to Figure ~\ref{fig:case}, LRMs have demonstrated intriguing capabilities on reasoning multi-hop erroneous code. Both Gemini 2.0 Flash Thinking and DeepSeek-R1 can promptly identify the \texttt{cause error line}, then mentally simulate code execution trace and explore multiple possible scenarios that could lead to runtime exception. However, Gemini 2.0 Flash Thinking was distracted by one of its early theories and produced an incorrect answer. On the other hand, DeepSeek-R1 ruled out all implausible possibilities after relentlessly pursuing an explanation for triggering a runtime error, eventually came up with the correct answer. 


\section{Related Work}

\paragraph{LLMs for Coding and Debugging}

LLM progress in code tasks led to benchmarks like HumanEval \citep{chen2021humaneval} and MBPP \citep{austin2021mbpp} for code synthesis, assessing syntactic correctness and functional accuracy. 
Runtime information is increasingly used in LLM debuggers \citep{zhong-etal-2024-debug-ldb}. Multiple benchmarks \citep{yang2025coastenhancingcodedebugging, tian-etal-2024-debugbench, jimenez2024swebenchlanguagemodelsresolve, ni2024nextteachinglargelanguage, gu2024cruxevalbenchmarkcodereasoning, jain2024livecodebenchholisticcontaminationfree, liu2024codemindframeworkchallengelarge} have focused on LLM debugging and code reasoning.

In data science coding, the landscape is evolving with general tools like the Data Interpreter \citep{hong2024datainterpreterllmagent} and specialized agents such as MatPlotAgent \citep{yang-etal-2024-matplotagent} and DSAgent \citep{guo2024dsagentautomateddatascience}. Benchmarks such as DSBench \citep{jing2024dsbenchfardatascience}, InfiAgent-DABench \citep{hu2024infiagentdabenchevaluatingagentsdata}, DSEval \citep{zhang2024benchmarkingdatascienceagents}, and PyBench \citep{zhang2024pybenchevaluatingllmagent} are emerging to evaluate the performance of LLMs in data science coding. However, they do not capture the complexities of real-world data science debugging.

DSDBench shifts focus to dynamic debugging of \textit{logical} errors, especially in complex data science workflows where runtime behavior and data dependencies are key.


\paragraph{LLM Self-Verification}

Self-correction enhance LLM reliability \citep{liang2024internalconsistencyselffeedbacklarge_survey}.  But, LLMs struggle to identify their own errors, especially in complex reasoning \citep{stechly2024selfverificationlimitationslargelanguage, tyen2024llmscannotfindreasoningerrorscorrect}. While some intrinsic self-correction exists \citep{liu2024largelanguagemodelsintrinsic}, its effectiveness for subtle logical errors is debated \citep{stechly2024selfverificationlimitationslargelanguage}.
Approaches to improve self-correction include confidence-guided methods \citep{li2024confidencemattersrevisitingintrinsic}, critique-focused training \citep{lin2024criticbenchbenchmarkingllmscritiquecorrect}.

However, self-verification research mainly targets general language tasks or simplified reasoning. DSDBench uniquely targets dynamic debugging of runtime errors in data science code.

\section{Conclusion}
We introduced DSDBench, a novel benchmark filling a critical gap in LLM evaluation by focusing on dynamic debugging of logical runtime errors in data science code, specifically multi-hop error tracing and multi-bug detection, built with a rigorous dataset construction process, reveals significant performance limitations of current state-of-the-art LLMs in these complex debugging scenarios.

\section*{Acknowledgments}
This research was supported by the Singapore Ministry of Education (MOE) Academic Research Fund (AcRF) Tier 1 grant (No. MSS24C004).

\section*{Limitations}
Our proposed DSDBench benchmark primarily focuses on the data science coding domain. While data science is a complex real-world task, our benchmark can be further expanded to encompass a wider range of practical coding scenarios, enabling a more comprehensive evaluation of LLMs' debugging performance in real-world coding pipelines. Additionally, future work could prioritize investigating LLMs' performance in debugging repository-level code with multi-file dependencies.

\section*{Ethical Considerations}

To construct the DSDBench benchmark, we employed human annotators for data labeling and verification tasks.  
We recruited annotators from our research institution holding at least a master degree in Computer Science. All annotators participated voluntarily and were provided with comprehensive information regarding the task's purpose, content, workload, and compensation prior to annotating.

\bibliography{custom}

\appendix

\section*{Appendix}

\begin{figure}[b]
\begin{tcolorbox}[colback=blue!2,colframe=blue!50!black]
\small
\textbf{SYSTEM PROMPT: } \texttt{You are a cutting-edge super capable code generation LLM. You will be given a natural language query, generate a runnable python code to satisfy all the requirements in the query. You can use any python library you want. When you complete a plot, remember to save it to a png file.}
\vspace{8pt} \\
\textbf{USER PROMPT: } \texttt{Here is the query:
"""
\{\{query\}\}
"""
If the query requires data manipulation from a csv file, process the data from the csv file and draw the plot in one piece of code. When you complete a plot, remember to save it to a png file. The file name should be """\{\{file\_name\}\}""".}
\end{tcolorbox}
\caption{The code generation prompt for code agent in Data Annotation.}
\label{fig:data_analysis_code_agent}
\end{figure}

\begin{figure}[htbp]
\begin{tcolorbox}[colback=blue!2,colframe=blue!50!black]
\small
\textbf{USER PROMPT: } \texttt{There are some errors in the code you gave:
\{\{error\_message\}\}
please correct the errors.
Then give the complete code and don't omit anything even though you have given it in the above code.}
\end{tcolorbox}
\caption{The self-debugging prompt for code agent in Data Annotation.}
\label{fig:Self_Debugging_prompt}
\end{figure}

\begin{figure}[ht]
\begin{tcolorbox}[colback=blue!2,colframe=blue!50!black]
\small
\texttt{You will be provided with an original query and a data analysis code. Your task is to:\\
1. Read the Question carefully, determine whether the code has followed the query requirements, if so, further identify any errors in its data analysis process. If the code faithfully followed seemingly wrong data analysis practices explicitly stated in the Question. Deem it as correct.\\
2. Explain any errors found, including:
   Explanation: Explain why this is an error and what issues it may cause.
   Expected Outcome: Explain how this error will affect the data analysis results, such as misleading outcomes, degraded performance, or incorrect interpretations.\\
Output Format:\\
json\\
    "is\_error": "true/false",\\
    "error\_explanation":\\
        "error\_type": "Describe the type of error",\\
        "explanation": "Detailed explanation of why this is an error and its impact",\\
        "expected\_outcome": "How this error will affect model performance or results",\\
        "suggestions": "Specific suggestions for fixing the error",\\
        "error\_type": "Another error type if multiple errors exist",\\
        "explanation": "Explanation for the second error",
        "expected\_outcome": "Expected outcome for the second error",\\
        "suggestions": "Suggestions for fixing the second error" \\     
Important Notes:\\
1. Always provide the output in the exact JSON format specified above\\
2. Set "is\_error" to "false" if no errors are found\\
3. If "is\_error" is "false", provide an empty array for error\_explanation\\
4. If "is\_error" is "true", include all identified errors in the error\_explanation array\\
5. Consider the original query requirements carefully, if the code follows the query's explicit requirements, even if they seem incorrect, consider it correct}
\end{tcolorbox}
\caption{The error verifying prompt in Data Annotation.}
\label{fig:error_verifying_prompt}
\end{figure}

\begin{figure}[htbp]
\begin{tcolorbox}[colback=blue!2,colframe=blue!50!black]
\small
\texttt{You will receive three components: \\
1. Original Query: A user query that contains specific requirements related to data analysis. \\
2. Correct Data Analysis Code: A working code snippet designed to analyze the data according to the original query. \\
3. CSV Information: Details about the structure content and sample data from the CSV file being analyzed. \\
Your task is to: \\
1. Identify sklearn and pandas code: Analyze the provided code and extract all lines where sklearn or pandas libraries are used. Organize these lines in a structured format. \\
2. Inject errors that will cause runtime interruptions: For EACH AND EVERY identified sklearn and pandas lines inject errors with the following guidelines: \\
Error Type: Inject errors that lead to runtime interruptions such as syntax errors attribute errors type errors or value errors. \\
Plausibility: The modified lines should still appear logical and plausible at first glance but contain mistakes that will cause the code to fail during execution. \\
Contextual alignment: Ensure the errors take into account the structure and content of the CSV file to create mistakes that are realistic and aligned with potential data issues. \\
Impact downstream processes: Errors should trigger runtime interruptions effectively halting the program before it completes execution. \\
3. Explain each error: For every injected error: \\
Describe why this is an error and the conditions under which it would fail. \\
Provide details on the likely runtime error e.g. KeyError ValueError AttributeError etc.. \\
4. Output the structured results: \\
Provide the original sklearn and pandas code in a structured list. \\
Include the complete modified code with runtimeinterrupting errors injected. \\
Clearly explain each injected error in a concise and structured format. \\
Return your output in the following JSON format: \\
original\_sklearn\_pandas\_code:  \\
Original sklearn or pandas code line \\
... \\
errors:  \\
code: Modified whole code file with the injected error \\
error\_type: Specify the type of runtimeinterrupting error e.g. KeyError ValueError etc. \\
explanation: Describe why this is an error and the conditions under which it will cause a runtime interruption
}
\end{tcolorbox}
\caption{The error injection prompt in Data Annotation.}
\label{fig:error_inject_prompt}
\end{figure}

\begin{figure}[ht]
\begin{tcolorbox}[colback=blue!2,colframe=blue!50!black]
\small
\textbf{SYSTEM PROMPT:} \texttt{You will be provided with an original query and a data analysis code. Your task is to: \\
1. Read the question carefully and identify if there are any logic error injected into the code. \\
2. For each logic error: \\
  - Locate the Cause: Specify the exact line of code that causes the issue. \\
  - Locate the Effect: Identify the line of code where the error will be triggered and the interpreter will throw an error. \\
  - Error Description: Provide a concise description of the error message thrown by the Python Interpreter (not the full traceback). \\
Output Format: \\
json  \\
     cause\_line: Specify the exact line of code causing the issue \\
     effect\_line: Specify the exact line of code where the error will be triggered \\
     error\_message: Provide a concise description of the error message thrown by the Python Interpreter not the full traceback 
 \\
There will be only one error in the code. Output only ONE json dict in your response.}
\end{tcolorbox}
\caption{The single error evaluation prompt for tested models.}
\label{fig:single_eval_prompt}
\end{figure}

\begin{figure}[t]
\begin{tcolorbox}[colback=blue!2,colframe=blue!50!black]
\small
\textbf{SYSTEM PROMPT:} \texttt{You will be provided with a data analysis code. Your task is to: \\
1. Read the code carefully and identify all logic errors injected into the code. There will be two or more logic errors in the code. \\
2. For each logic error you identify: \\
  - Locate the Cause: Specify the exact line of code that causes the issue. \\
  - Locate the Effect: Identify the line of code where the error will be triggered and the interpreter will throw an error or where the incorrect behavior is observed. \\
  - Error Description: Provide a concise description of the error message thrown by the Python Interpreter not the full traceback. Focus on the type of error and the reason if possible from the output. \\
Output Format: \\
json \\
     cause\_line: Specify the exact line of code causing error 1 \\
     effect\_line: Specify the exact line of code where error 1 is triggered \\
     error\_message: Concise error message for error 1 
        cause\_line: Specify the exact line of code causing error 2 \\
     effect\_line: Specify the exact line of code where error 2 is triggered \\
     error\_message: Concise error message for error 2 
    ... and so on for all identified errors 
There will be more than one error in the code. BUT output only ONE json block in your response.}
\end{tcolorbox}
\caption{The multi error evaluation prompt for tested models.}
\label{fig:multi_eval_prompt}
\end{figure}

\section{Data Annotation Agent}
\label{app: data_annotation}
Our automatic data annotation agent is comprised of two components, a self-debugging code agent and an error verifier agent. The prompts used for these agents are in Figure \ref{fig:data_analysis_code_agent}, \ref{fig:Self_Debugging_prompt}, \ref{fig:error_verifying_prompt}.

the code agent receives benchmark questions as input, generate a draft code according to the requirements in the questions. Then, the system environment in which the agent framework operates executes the draft code. If not successfully executed, the interpreter error message will be passed to the self-debugging code agent, prompting the agent to generate another draft code according to the error message and original benchmark question. The agent will be given a set amount of chances to refine its code according to the error message, if the code is still not executable after 5 rounds, the agent stops. If the code successfully executed within 5 retry times, then the error verifier agent will step in and check the code for further logical errors that may not elicit an interpreter error. If the error verifier agent deems the code correct, the system environment will execute the code and extract the answers from the code. Then we will compare the model generated answers with ground truth answers in each benchmark, if the answers match, we will collect the code that produces these answers as the correct code for our subsequent annotation process.

\section{Prompts for Error Injection}
\label{app:error_injection}
Figure \ref{fig:error_inject_prompt} demonstrates the prompt for error injection, the LLM injector is required to inject plausible runtime logical error into existing correct code with meta information such as benchmark question, data file information. The output format should be a well-formatted JSON dict.

\section{Error Types}
\label{app:error_type}
The error types collected in our benchmark are all Python Built-in Exceptions, more information can be accessed at: \url{https://docs.python.org/3/library/exceptions.html}

\section{Prompts for Evaluation}
\label{app:evaluation_prompt}
Figure \ref{fig:single_eval_prompt} and \ref{fig:multi_eval_prompt} demonstrates the prompts used for evaluating LLMs and LRMs on single bug and multi bug detection. The models are provided with a benchmark question and a snippet of buggy code. The models should identify the error and locate cause and effect error line of code and reproduce error message thrown by the Python Interpreter. The output for single bug detection should be a well-formatted JSON dict, the output for multi bug detection should a list of aforementioned JSON dict.

\section{Full Evaluation Results}
\label{app:full_metrics}
We provide the full results of Single-Bug and Multi-Bug evaluation with all four metrics in Table \ref{tab:llm_single_full}, \ref{tab:llm_multi_full}, \ref{tab:lrm_single_full} and \ref{tab:lrm_multi_full}.

\begin{table*}[b]
    \centering
    \footnotesize

    \begin{adjustbox}{width=\textwidth}
    \begin{tabular}{l|*{4}{S}|*{4}{S}|*{4}{S}|*{4}{S}} 
        \toprule
         \multirow{2}{*}{\textbf{Model}} & \multicolumn{4}{c|}{\textbf{Cause Line}} & \multicolumn{4}{c|}{\textbf{Effect Line}} & \multicolumn{4}{c|}{\textbf{Error Type}} & \multicolumn{4}{c}{\textbf{Error Message}} \\
         & {\textbf{P}} & {\textbf{R}} & {\textbf{F1}} & {\textbf{Acc}} & {\textbf{P}} & {\textbf{R}} & {\textbf{F1}} & {\textbf{Acc}} & {\textbf{P}} & {\textbf{R}} & {\textbf{F1}} & {\textbf{Acc}} & {\textbf{P}} & {\textbf{R}} & {\textbf{F1}} & {\textbf{Acc}} \\
        \midrule
        gpt-4o & \num{39.5} & \num{39.0} & \num{39.2} & \num{39.0} & \num{34.7} & \num{34.3} & \num{34.5} & \num{34.3} & \num{31.0} & \num{30.6} & \num{30.8} & \num{30.6} & \num{31.8} & \num{31.4} & \num{31.6} & \num{31.4} \\
        gpt-4o-mini & \num{43.3} & \num{40.2} & \num{41.7} & \num{40.2} & \num{25.7} & \num{23.9} & \num{24.8} & \num{23.9} & \num{23.4} & \num{21.7} & \num{22.5} & \num{21.7} & \num{23.0} & \num{21.3} & \num{22.1} & \num{21.3} \\
        claude-3-5-sonnet & \num{45.4} & \num{43.7} & \num{44.6} & \num{43.7} & \num{36.6} & \num{35.2} & \num{35.9} & \num{35.2} & \num{37.7} & \num{36.3} & \num{37.0} & \num{36.3} & \num{35.3} & \num{34.0} & \num{34.7} & \num{34.0} \\
        llama-3.1-8b-instant & \num{32.4} & \num{25.2} & \num{28.4} & \num{25.2} & \num{18.2} & \num{14.2} & \num{15.9} & \num{14.2} & \num{9.9} & \num{7.7} & \num{8.6} & \num{7.7} & \num{9.2} & \num{7.2} & \num{8.0} & \num{7.2} \\
        llama-3.1-70b-versatile & \num{45.7} & \num{42.5} & \num{44.0} & \num{42.5} & \num{31.4} & \num{29.3} & \num{30.3} & \num{29.3} & \num{21.9} & \num{20.4} & \num{21.1} & \num{20.4} & \num{22.5} & \num{20.9} & \num{21.7} & \num{20.9} \\
        llama-3.1-405b-instruct & \num{46.9} & \num{41.7} & \num{44.1} & \num{41.7} & \num{35.2} & \num{31.3} & \num{33.1} & \num{31.3} & \num{32.9} & \num{29.3} & \num{31.0} & \num{29.3} & \num{32.9} & \num{29.3} & \num{31.0} & \num{29.3} \\
        Qwen2.5-7B-Instruct & \num{31.0} & \num{29.3} & \num{30.1} & \num{29.3} & \num{20.4} & \num{19.3} & \num{19.8} & \num{19.3} & \num{11.3} & \num{10.7} & \num{11.0} & \num{10.7} & \num{11.6} & \num{10.9} & \num{11.2} & \num{10.9} \\
        Qwen2.5-32B-Instruct & \num{43.5} & \num{40.9} & \num{42.1} & \num{40.9} & \num{32.4} & \num{30.5} & \num{31.4} & \num{30.5} & \num{26.3} & \num{24.7} & \num{25.5} & \num{24.7} & \num{26.3} & \num{24.7} & \num{25.5} & \num{24.7} \\
        Qwen2.5-72B-Instruct & \num{43.8} & \num{41.6} & \num{42.6} & \num{41.6} & \num{38.1} & \num{36.2} & \num{37.1} & \num{36.2} & \num{29.0} & \num{27.5} & \num{28.2} & \num{27.5} & \num{28.8} & \num{27.4} & \num{28.1} & \num{27.4} \\
        deepseek-chat & \num{50.6} & \num{48.3} & \num{49.4} & \num{48.3} & \num{36.2} & \num{34.5} & \num{35.4} & \num{34.5} & \num{37.6} & \num{35.9} & \num{36.7} & \num{35.9} & \num{36.4} & \num{34.7} & \num{35.5} & \num{34.7} \\
        \bottomrule
    \end{tabular}
    \end{adjustbox}
    \caption{Overall evaluation results of Single-Bug Detection on DSDBench. P=Precision, R=Recall, F1=F1-Score, Acc=Accuracy.}
    \label{tab:llm_single_full}
\end{table*}

\begin{table*}[b]
    \centering
    \footnotesize

    \begin{adjustbox}{width=\textwidth}
    \begin{tabular}{l|*{4}{S}|*{4}{S}|*{4}{S}|*{4}{S}}
        \toprule
         \multirow{2}{*}{\textbf{Model}} & \multicolumn{4}{c|}{\textbf{Cause Line}} & \multicolumn{4}{c|}{\textbf{Effect Line}} & \multicolumn{4}{c|}{\textbf{Error Type}} & \multicolumn{4}{c}{\textbf{Error Message}} \\
         & {\textbf{P}} & {\textbf{R}} & {\textbf{F1}} & {\textbf{Acc}} & {\textbf{P}} & {\textbf{R}} & {\textbf{F1}} & {\textbf{Acc}} & {\textbf{P}} & {\textbf{R}} & {\textbf{F1}} & {\textbf{Acc}} & {\textbf{P}} & {\textbf{R}} & {\textbf{F1}} & {\textbf{Acc}} \\
        \midrule
        gpt-4o & \num{20.5} & \num{20.3} & \num{20.4} & \num{20.3} & \num{10.5} & \num{10.4} & \num{10.5} & \num{10.4} & \num{3.6} & \num{3.6} & \num{3.6} & \num{3.6} & \num{4.7} & \num{4.7} & \num{4.7} & \num{4.7} \\
        gpt-4o-mini & \num{11.3} & \num{11.2} & \num{11.2} & \num{11.2} & \num{2.7} & \num{2.7} & \num{2.7} & \num{2.7} & \num{2.2} & \num{2.2} & \num{2.2} & \num{2.2} & \num{0.8} & \num{0.8} & \num{0.8} & \num{0.8} \\
        claude-3-5-sonnet & \num{12.5} & \num{12.3} & \num{12.4} & \num{12.3} & \num{4.2} & \num{4.1} & \num{4.1} & \num{4.1} & \num{1.9} & \num{1.9} & \num{1.9} & \num{1.9} & \num{2.5} & \num{2.5} & \num{2.5} & \num{2.5} \\
        llama-3.1-8b-instant & \num{5.1} & \num{3.0} & \num{3.8} & \num{3.0} & \num{0.0} & \num{0.0} & \num{0.0} & \num{0.0} & \num{0.0} & \num{0.0} & \num{0.0} & \num{0.0} & \num{0.0} & \num{0.0} & \num{0.0} & \num{0.0} \\
        llama-3.1-70b-versatile & \num{0.0} & \num{0.0} & \num{0.0} & \num{0.0} & \num{0.0} & \num{0.0} & \num{0.0} & \num{0.0} & \num{0.0} & \num{0.0} & \num{0.0} & \num{0.0} & \num{0.0} & \num{0.0} & \num{0.0} & \num{0.0} \\
        llama-3.1-405b-instruct & \num{24.2} & \num{18.6} & \num{21.1} & \num{18.6} & \num{11.0} & \num{8.5} & \num{9.6} & \num{8.5} & \num{1.4} & \num{1.1} & \num{1.2} & \num{1.1} & \num{3.2} & \num{2.5} & \num{2.8} & \num{2.5} \\
        Qwen2.5-7B-Instruct & \num{5.9} & \num{4.7} & \num{5.2} & \num{4.7} & \num{1.4} & \num{1.1} & \num{1.2} & \num{1.1} & \num{0.3} & \num{0.3} & \num{0.3} & \num{0.3} & \num{0.0} & \num{0.0} & \num{0.0} & \num{0.0} \\
        Qwen2.5-32B-Instruct & \num{17.6} & \num{17.5} & \num{17.6} & \num{17.5} & \num{6.3} & \num{6.3} & \num{6.3} & \num{6.3} & \num{2.2} & \num{2.2} & \num{2.2} & \num{2.2} & \num{2.2} & \num{2.2} & \num{2.2} & \num{2.2} \\
        Qwen2.5-72B-Instruct & \num{21.4} & \num{21.4} & \num{21.4} & \num{21.4} & \num{11.2} & \num{11.2} & \num{11.2} & \num{11.2} & \num{3.0} & \num{3.0} & \num{3.0} & \num{3.0} & \num{3.6} & \num{3.6} & \num{3.6} & \num{3.6} \\
        deepseek-chat & \num{15.2} & \num{15.1} & \num{15.1} & \num{15.1} & \num{6.6} & \num{6.6} & \num{6.6} & \num{6.6} & \num{3.3} & \num{3.3} & \num{3.3} & \num{3.3} & \num{4.7} & \num{4.7} & \num{4.7} & \num{4.7} \\
        \bottomrule
    \end{tabular}
    \end{adjustbox}
    \caption{Overall evaluation results of Multi-Bug Detection on DSDBench. P=Precision, R=Recall, F1=F1-Score, Acc=Accuracy.}
    \label{tab:llm_multi_full}
\end{table*}

\begin{table*}[t]
    \centering
    \footnotesize

    \begin{adjustbox}{width=\textwidth}
    \begin{tabular}{l|*{4}{S}|*{4}{S}|*{4}{S}|*{4}{S}}
        \toprule
         \multirow{2}{*}{\textbf{Model}} & \multicolumn{4}{c|}{\textbf{Cause Line}} & \multicolumn{4}{c|}{\textbf{Effect Line}} & \multicolumn{4}{c|}{\textbf{Error Type}} & \multicolumn{4}{c}{\textbf{Error Message}} \\
         & {\textbf{P}} & {\textbf{R}} & {\textbf{F1}} & {\textbf{Acc}} & {\textbf{P}} & {\textbf{R}} & {\textbf{F1}} & {\textbf{Acc}} & {\textbf{P}} & {\textbf{R}} & {\textbf{F1}} & {\textbf{Acc}} & {\textbf{P}} & {\textbf{R}} & {\textbf{F1}} & {\textbf{Acc}} \\
        \midrule
        gpt-4o & \num{35.8} & \num{35.4} & \num{35.6} & \num{35.4} & \num{31.6} & \num{31.2} & \num{31.4} & \num{31.2} & \num{33.7} & \num{33.3} & \num{33.5} & \num{33.3} & \num{33.7} & \num{33.3} & \num{33.5} & \num{33.3} \\
        gpt-4o-mini & \num{42.7} & \num{39.6} & \num{41.1} & \num{39.6} & \num{31.5} & \num{29.2} & \num{30.3} & \num{29.2} & \num{27.0} & \num{25.0} & \num{25.9} & \num{25.0} & \num{24.7} & \num{22.9} & \num{23.8} & \num{22.9} \\
        claude-3-5-sonnet & \num{37.0} & \num{35.4} & \num{36.2} & \num{35.4} & \num{27.2} & \num{26.0} & \num{26.6} & \num{26.0} & \num{34.8} & \num{33.3} & \num{34.0} & \num{33.3} & \num{32.6} & \num{31.2} & \num{31.9} & \num{31.2} \\
        llama-3.1-8b-instant & \num{24.1} & \num{13.5} & \num{17.3} & \num{13.5} & \num{20.4} & \num{11.5} & \num{14.7} & \num{11.5} & \num{11.1} & \num{6.2} & \num{8.0} & \num{6.2} & \num{9.3} & \num{5.2} & \num{6.7} & \num{5.2} \\
        llama-3.1-70b-versatile & \num{36.7} & \num{34.4} & \num{35.5} & \num{34.4} & \num{23.3} & \num{21.9} & \num{22.6} & \num{21.9} & \num{20.0} & \num{18.8} & \num{19.4} & \num{18.8} & \num{20.0} & \num{18.8} & \num{19.4} & \num{18.8} \\
        llama-3.1-405b-instruct & \num{51.2} & \num{43.8} & \num{47.2} & \num{43.8} & \num{37.8} & \num{32.3} & \num{34.8} & \num{32.3} & \num{36.6} & \num{31.2} & \num{33.7} & \num{31.2} & \num{40.2} & \num{34.4} & \num{37.1} & \num{34.4} \\
        Qwen2.5-7B-Instruct & \num{30.8} & \num{29.2} & \num{29.9} & \num{29.2} & \num{24.2} & \num{22.9} & \num{23.5} & \num{22.9} & \num{12.1} & \num{11.5} & \num{11.8} & \num{11.5} & \num{13.2} & \num{12.5} & \num{12.8} & \num{12.5} \\
        Qwen2.5-32B-Instruct & \num{35.2} & \num{32.3} & \num{33.7} & \num{32.3} & \num{28.4} & \num{26.0} & \num{27.2} & \num{26.0} & \num{33.0} & \num{30.2} & \num{31.5} & \num{30.2} & \num{26.1} & \num{24.0} & \num{25.0} & \num{24.0} \\
        Qwen2.5-72B-Instruct & \num{26.7} & \num{25.0} & \num{25.8} & \num{25.0} & \num{32.2} & \num{30.2} & \num{31.2} & \num{30.2} & \num{30.0} & \num{28.1} & \num{29.0} & \num{28.1} & \num{27.8} & \num{26.0} & \num{26.9} & \num{26.0} \\
        deepseek-chat & \num{49.4} & \num{44.8} & \num{47.0} & \num{44.8} & \num{31.0} & \num{28.1} & \num{29.5} & \num{28.1} & \num{37.9} & \num{34.4} & \num{36.1} & \num{34.4} & \num{37.9} & \num{34.4} & \num{36.1} & \num{34.4} \\
        gemini-2.0-flash & \num{49.4} & \num{42.7} & \num{45.8} & \num{42.7} & \num{37.3} & \num{32.3} & \num{34.6} & \num{32.3} & \num{38.6} & \num{33.3} & \num{35.8} & \num{33.3} & \num{41.0} & \num{35.4} & \num{38.0} & \num{35.4} \\
        deepseek-r1 & \num{51.6} & \num{49.0} & \num{50.3} & \num{49.0} & \num{51.6} & \num{49.0} & \num{50.3} & \num{49.0} & \num{56.0} & \num{53.1} & \num{54.5} & \num{53.1} & \num{57.1} & \num{54.2} & \num{55.6} & \num{54.2} \\
        o1-mini & \num{46.2} & \num{43.8} & \num{44.9} & \num{43.8} & \num{38.5} & \num{36.5} & \num{37.4} & \num{36.5} & \num{46.2} & \num{43.8} & \num{44.9} & \num{43.8} & \num{49.5} & \num{46.9} & \num{48.1} & \num{46.9} \\
        \bottomrule
    \end{tabular}
    \end{adjustbox}
    \caption{Comparison with large reasoning models (LRMs) on Single-Bug Detection. P=Precision, R=Recall, F1=F1-Score, Acc=Accuracy.}
    \label{tab:lrm_single_full}
\end{table*}

\begin{table*}[t]
    \centering
    \footnotesize

    \begin{adjustbox}{width=\textwidth}
    \begin{tabular}{l|*{4}{S}|*{4}{S}|*{4}{S}|*{4}{S}}
        \toprule
         \multirow{2}{*}{\textbf{Model}} & \multicolumn{4}{c|}{\textbf{Cause Line}} & \multicolumn{4}{c|}{\textbf{Effect Line}} & \multicolumn{4}{c|}{\textbf{Error Type}} & \multicolumn{4}{c}{\textbf{Error Message}} \\
         & {\textbf{P}} & {\textbf{R}} & {\textbf{F1}} & {\textbf{Acc}} & {\textbf{P}} & {\textbf{R}} & {\textbf{F1}} & {\textbf{Acc}} & {\textbf{P}} & {\textbf{R}} & {\textbf{F1}} & {\textbf{Acc}} & {\textbf{P}} & {\textbf{R}} & {\textbf{F1}} & {\textbf{Acc}} \\
        \midrule
        gpt-4o & \num{12.8} & \num{12.5} & \num{12.7} & \num{12.5} & \num{5.1} & \num{5.0} & \num{5.1} & \num{5.0} & \num{2.6} & \num{2.5} & \num{2.5} & \num{2.5} & \num{2.6} & \num{2.5} & \num{2.5} & \num{2.5} \\
        gpt-4o-mini & \num{7.5} & \num{7.5} & \num{7.5} & \num{7.5} & \num{5.0} & \num{5.0} & \num{5.0} & \num{5.0} & \num{2.5} & \num{2.5} & \num{2.5} & \num{2.5} & \num{0.0} & \num{0.0} & \num{0.0} & \num{0.0} \\
        claude-3-5-sonnet & \num{10.3} & \num{10.0} & \num{10.1} & \num{10.0} & \num{7.7} & \num{7.5} & \num{7.6} & \num{7.5} & \num{5.1} & \num{5.0} & \num{5.1} & \num{5.0} & \num{7.7} & \num{7.5} & \num{7.6} & \num{7.5} \\
        llama-3.1-8b-instant & \num{0.0} & \num{0.0} & \num{0.0} & \num{0.0} & \num{0.0} & \num{0.0} & \num{0.0} & \num{0.0} & \num{0.0} & \num{0.0} & \num{0.0} & \num{0.0} & \num{0.0} & \num{0.0} & \num{0.0} & \num{0.0} \\
        llama-3.1-70b-versatile & \num{0.0} & \num{0.0} & \num{0.0} & \num{0.0} & \num{0.0} & \num{0.0} & \num{0.0} & \num{0.0} & \num{0.0} & \num{0.0} & \num{0.0} & \num{0.0} & \num{0.0} & \num{0.0} & \num{0.0} & \num{0.0} \\
        llama-3.1-405b-instruct & \num{23.3} & \num{17.5} & \num{20.0} & \num{17.5} & \num{16.7} & \num{12.5} & \num{14.3} & \num{12.5} & \num{6.7} & \num{5.0} & \num{5.7} & \num{5.0} & \num{6.7} & \num{5.0} & \num{5.7} & \num{5.0} \\
        Qwen2.5-7B-Instruct & \num{3.3} & \num{2.5} & \num{2.9} & \num{2.5} & \num{0.0} & \num{0.0} & \num{0.0} & \num{0.0} & \num{0.0} & \num{0.0} & \num{0.0} & \num{0.0} & \num{0.0} & \num{0.0} & \num{0.0} & \num{0.0} \\
        Qwen2.5-32B-Instruct & \num{17.5} & \num{17.5} & \num{17.5} & \num{17.5} & \num{0.0} & \num{0.0} & \num{0.0} & \num{0.0} & \num{5.0} & \num{5.0} & \num{5.0} & \num{5.0} & \num{2.5} & \num{2.5} & \num{2.5} & \num{2.5} \\
        Qwen2.5-72B-Instruct & \num{22.5} & \num{22.5} & \num{22.5} & \num{22.5} & \num{17.5} & \num{17.5} & \num{17.5} & \num{17.5} & \num{2.5} & \num{2.5} & \num{2.5} & \num{2.5} & \num{5.0} & \num{5.0} & \num{5.0} & \num{5.0} \\
        deepseek-chat & \num{12.8} & \num{12.5} & \num{12.7} & \num{12.5} & \num{7.7} & \num{7.5} & \num{7.6} & \num{7.5} & \num{5.1} & \num{5.0} & \num{5.1} & \num{5.0} & \num{7.7} & \num{7.5} & \num{7.6} & \num{7.5} \\
        o1-mini & \num{37.8} & \num{35.0} & \num{36.4} & \num{35.0} & \num{24.3} & \num{22.5} & \num{23.4} & \num{22.5} & \num{18.9} & \num{17.5} & \num{18.2} & \num{17.5} & \num{18.9} & \num{17.5} & \num{18.2} & \num{17.5} \\
        gemini-2.0-flash & \num{21.1} & \num{20.0} & \num{20.5} & \num{20.0} & \num{13.2} & \num{12.5} & \num{12.8} & \num{12.5} & \num{0.0} & \num{0.0} & \num{0.0} & \num{0.0} & \num{2.6} & \num{2.5} & \num{2.6} & \num{2.5} \\
        deepseek-r1 & \num{32.5} & \num{32.5} & \num{32.5} & \num{32.5} & \num{25.0} & \num{25.0} & \num{25.0} & \num{25.0} & \num{15.0} & \num{15.0} & \num{15.0} & \num{15.0} & \num{15.0} & \num{15.0} & \num{15.0} & \num{15.0} \\
        \bottomrule
    \end{tabular}
    \end{adjustbox}
    \caption{Comparison with large reasoning models (LRMs) on Multi-Bug Detection. P=Precision, R=Recall, F1=F1-Score, Acc=Accuracy.}
    \label{tab:lrm_multi_full}
\end{table*}

\section{Prompting Strategies and Test-Time Compute}
\label{app:prompting_strategies}

We evaluate four test-time reasoning strategies beyond direct zero-shot prompting on the \emph{single-bug subset} used for LRMs: \emph{Zero-shot CoT}, \emph{One-shot CoT}, \emph{Self-Refine}, and \emph{Self-Consistency}. All runs disallow external tools and execution to isolate in-context reasoning. See results in Table \ref{tab:prompting_strategies}.

\begin{table*}[t]
\centering
\resizebox{\textwidth}{!}{
\begin{tabular}{l l c c c c}
\toprule
\textbf{Model} & \textbf{Prompting} & \textbf{Cause Acc.} & \textbf{Effect Acc.} & \textbf{$\Delta$ Cause vs. Direct} & \textbf{$\Delta$ Effect vs. Direct} \\
\midrule
GPT-4o & Direct (Baseline) & 35.4 & 31.2 & -- & -- \\
GPT-4o & Zero-shot CoT & 33.3 & 30.2 & -2.1 & -1.0 \\
GPT-4o & One-shot CoT & 33.3 & 31.3 & -2.1 & +0.1 \\
GPT-4o & Self-Refine & 33.3 & 30.2 & -2.1 & -1.0 \\
GPT-4o & Self-Consistency & 37.5 & 34.4 & +2.1 & +3.2 \\
\midrule
DeepSeek-V3 & Direct (Baseline) & 44.8 & 28.1 & -- & -- \\
DeepSeek-V3 & Zero-shot CoT & 44.8 & 33.3 & +0.0 & +5.2 \\
DeepSeek-V3 & One-shot CoT & 46.9 & 33.3 & +2.1 & +5.2 \\
DeepSeek-V3 & Self-Refine & 47.9 & 31.3 & +3.1 & +3.2 \\
DeepSeek-V3 & Self-Consistency & 42.7 & 37.5 & -2.1 & +9.4 \\
\midrule
Qwen2.5-72B & Direct (Baseline) & 26.0 & 26.0 & -- & -- \\
Qwen2.5-72B & Zero-shot CoT & 37.5 & 31.3 & +11.5 & +5.3 \\
Qwen2.5-72B & One-shot CoT & 37.5 & 25.0 & +11.5 & -1.0 \\
Qwen2.5-72B & Self-Refine & 37.5 & 31.3 & +11.5 & +5.3 \\
Qwen2.5-72B & Self-Consistency & 40.6 & 36.5 & +14.6 & +10.5 \\
\bottomrule
\end{tabular}}
\caption{Prompting strategies on the \emph{single-bug subset}. Gains show sensitivity to test-time compute and scaffolding; however, all models remain far from saturation.}
\label{tab:prompting_strategies}
\end{table*}

\textbf{Observations.} Self-Consistency is most helpful for GPT-4o; DeepSeek-V3 benefits more on effect-line localization; Qwen2.5-72B experiences large gains, reflecting stronger utility from added scaffolds. Despite improvements, the absolute accuracies remain modest, underscoring DSDBench’s difficulty.

\section{Agentic Systems: Pilot Evaluation}
\label{app:agentic_pilot}

We evaluate whether agentic workflows mitigate DSDBench’s challenges using Cursor Agent (Claude 3.5 Sonnet backend) and OpenHands on the \emph{single-bug subset}. We consider two settings:
\begin{itemize}[leftmargin=*,nosep]
\item \textbf{Single-Attempt Repair (pass@1):} Canonicalize each item into a one-line repair task; the agent may change \emph{exactly one line} once. Success requires a clean reference run.
\item \textbf{Unconstrained Agentic Workflow:} The agent may iteratively invoke tools and execute code until success (no edit budget).
\end{itemize}
See results in Table \ref{tab:agentic_pilot_table}

\begin{table*}[t]
\centering
\resizebox{\textwidth}{!}{
\begin{tabular}{l l l c l}
\toprule
\textbf{Model / Approach} & \textbf{Task} & \textbf{Metric} & \textbf{Performance (\%)} & \textbf{Key Insight} \\
\midrule
Claude-3.5-Sonnet (Standalone) & Cause Line Localization & Accuracy & 34.0 & Baseline “understanding” is low. \\
Cursor Agent (Claude-3.5-Sonnet) & Single-Attempt Repair & pass@1 & 48.9 & Recasting to generation yields +14.9 points. \\
Cursor Agent (Claude-3.5-Sonnet) & Unconstrained Repair & Pass Rate & $\sim$100 & Often succeeds via broad, non-minimal rewrites. \\
OpenHands (Cloud GUI)$^{\dagger}$ & Single-Attempt Repair & pass@1 & 18.6 & Large variance across agents/backbones. \\
OpenHands (Cloud GUI)$^{\dagger}$ & Unconstrained Repair & Pass Rate & 41.7 & Iterative repair remains non-trivial. \\
\bottomrule
\end{tabular}}
\caption{Agentic pilot on the \emph{single-bug subset}. $^{\dagger}$Backbone LLM unknown for the cloud GUI; results are indicative only.}
\label{tab:agentic_pilot_table}
\end{table*}

\textbf{Takeaways.} (i) Single-attempt agentic repair outperforms standalone localization, indicating that reframing from diagnosis to generation aligns better with autoregressive training. (ii) Unconstrained agents can brute-force success by extensive regeneration, masking the \emph{reasoning} gap DSDBench is designed to expose. This justifies our focus on line-level localization and semantic message reproduction as complementary to patch-level pass rates.

\end{document}